\pdfoutput=1
\documentclass[11pt]{article}

\usepackage[]{arxiv}

\usepackage{times}
\usepackage{latexsym}

\usepackage{microtype}
\usepackage{caption,subcaption}
\usepackage{booktabs} % for professional tables
\usepackage{textcomp}
\usepackage{hyperref}
\usepackage{graphicx} % DO NOT CHANGE THIS
\usepackage{natbib}  % DO NOT CHANGE THIS AND DO NOT ADD ANY OPTIONS TO IT
\usepackage{caption} % DO NOT CHANGE THIS AND DO NOT ADD ANY OPTIONS TO IT
\usepackage{multicol}
\usepackage{multirow}

\usepackage{tikz}
\usetikzlibrary{bayesnet}
\usepackage{amsmath,amsfonts,amssymb,amsthm}
\usepackage{soul}
\usepackage{algorithm}
\usepackage[noend]{algpseudocode}
\usepackage{booktabs}
\usepackage[switch]{lineno}  %
\usepackage{mdwlist}
\usepackage{enumitem}
\newcommand{\bs}[1]{\boldsymbol{#1}}

\usepackage[T1]{fontenc}
\usepackage[utf8]{inputenc}

\usepackage{microtype}

\title{Bayesian Topic Regression for Causal Inference}

\author{Maximilian Ahrens\textsuperscript{1}, Julian Ashwin\textsuperscript{1}, Jan-Peter Calliess\textsuperscript{1}, Vu Nguyen\textsuperscript{1,2} \\
  \textsuperscript{1}University of Oxford, \textsuperscript{2}Amazon \\
  \texttt{\{mahrens,jan\}@robots.ox.ac.uk}, \texttt{julian.ashwin@economics.ox.ac.uk}, \\
  \texttt{vutngn@amazon.com} \\
  }

\begin{document}

\maketitle 
%-------------------------------------------------------------
%-------------------------------------------------------------
%-------------------------------------------------------------
\begin{abstract}
Causal inference using observational text data is becoming increasingly popular in many research areas.  This paper presents the Bayesian Topic Regression (BTR) model that uses both text and numerical information to model an outcome variable. It allows estimation of both discrete and continuous treatment effects. Furthermore, it allows for the inclusion of additional numerical confounding factors next to text data. To this end, we combine a supervised Bayesian topic model with a Bayesian regression framework and perform supervised representation learning for the text features jointly with the regression parameter training, respecting the Frisch-Waugh-Lovell theorem.
Our paper makes two main contributions. First, we provide a regression framework that allows causal inference in settings when both text and numerical confounders are of relevance. We show with synthetic and semi-synthetic datasets that our joint approach recovers ground truth with lower bias than any benchmark model, when text and numerical features are correlated.
Second, experiments on two real-world datasets demonstrate that a joint and supervised learning strategy also yields superior prediction results compared to strategies that estimate regression weights for text and non-text features separately, being even competitive with more complex deep neural networks.
\end{abstract}
%-------------------------------------------------------------
%-------------------------------------------------------------
%-------------------------------------------------------------

\section{Introduction}
Causal inference using observational text data is increasingly popular across many research areas \citep{keith2020text}. It expands the range of research questions that can be explored when using text data across various fields, such as in the social and data sciences; adding to an extensive literature of text analysis methods and applications \citet{grimmer2013text_as_data,gentzkow2019text}.
Where randomized controlled trials are not possible, observational data might often be the only source of information and statistical methods need to be deployed to adjust for confounding biases. Text data can either serve as a proxy for otherwise unobserved confounding variables, be a confounding factor in itself, or even represent the treatment or outcome variable of interest. 

\textbf{The framework:}
We consider the causal inference settings where we allow for the treatment variable to be binary, categorical or continuous. In our setting, text might be either a confounding factor or a proxy for a latent confounding variable. We also allow for additional non-text confounders (covariates). To the best of our knowledge, we are the first to provide such statistical inference framework. 

Considering both text and numerical data jointly can not only improve prediction performance, but can be crucial for conducting unbiased statistical inference. When treatment and confounders are correlated with each other and with the outcome, the Frisch-Waugh-Lovell theorem \cite{frisch1933partial,lovell1963seasonal}, described in Section \ref{section:CEF}, implies that all regression weights must be estimated jointly, otherwise estimates will be biased. Text features themselves are `estimated data'. If they stem from supervised learning, which estimated the text features with respect to the outcome variable separately from the numerical features, then the resulting estimated (causal) effects will be biased.

\textbf{Our contributions:} With this paper, we introduce a Bayesian Topic Regression (BTR) framework that combines a Bayesian topic model with a Bayesian regression approach. This allows us to perform supervised representation learning for text features jointly with the estimation of regression parameters that include both treatment and additional numerical covariates. In particular, information about dependencies between outcome, treatment and controls does not only inform the regression part, but directly feeds into the topic modelling process.
Our approach aims towards estimating `causally sufficient' text representations in the spirit of \citet{veitch2020adapting}. We  
show on both synthetic and semi-synthetic datasets that our BTR model recovers the ground truth more accurately than a wide range of benchmark models. 
Finally, we demonstrate on two real-world customer review datasets - \emph{Yelp} and \emph{Booking.com} - that a joint supervised learning strategy, using both text and non-text features, also improves prediction accuracy of the target variable compared to a `two-step' estimation approach with the same models. This does not come at a cost of higher perplexity scores on the document modelling task. We also show that relatively simple supervised topic models with a linear regression layer that follow such joint approach can even compete with much more complex, non-linear deep neural networks that do not follow the joint estimation approach.

%-------------------------------------------------------------
%-------------------------------------------------------------
%-------------------------------------------------------------
\section{Background and Related Work}\label{section:background}
\subsection{Causal Inference with Text}
\citet{egami2018make}  and \citet{wood2018challenges} provide a comprehensive conceptional framework for inference  with text and outline the challenges, focusing on text as treatment and outcome. In a similar vein, \citet{tan_2014,fong_grimmer_2016} focus on text as treatment. \citet{roberts2020adjusting, mozer2020} address adjustment for text as a confounder via text matching considering both topic and word level features. 
\citet{veitch2020adapting} introduce a framework to estimate causally sufficient text representations via topic and general language models. Like us, they consider text as a confounder. Their framework exclusively focuses on binary treatment effects and does not allow for additional numerical confounders. We extend this framework.\\
\noindent \textbf{Causal inference framework with text:}
This general framework hinges on the assumption that through supervised dimensionality reduction of the text, we can identify text representations that capture the correlations with the outcome, the treatment and other control variables. Assume we observe iid data tuples $D_i = (y_i, r_i, \bs{W_i}, \bs{C_i})$, where for observation $i$, $y_i$ is the outcome, $t_i$ is the treatment, $\bs {W_i}$ is the associated text, and $\bs{C_i}$ are other confounding effects for which we have numerical measurements. Following the notational conventions set out in \cite{pearl_2009causality}, define the average treatment effect of the treated (ATT\footnote{depicted is the ATT of a binary treatment. The same logic applies for categorical or continuous treatments.}) as:
\begin{equation*}
\small
\begin{split}
\delta = \mathbb{E}[ y |\text{do}( t =1), t =1]-\mathbb{E}[ y | \text{do}( t =0), t =1].
\end{split}
\end{equation*}
In the spirit of \citet{veitch2020adapting}, we assume that our models can learn a supervised text representation $\bs{Z_i} = g(\bs{W_i},y_i,t_i,\bs{C_i})$, which in our case, together with $\bs{C_i}$ blocks all `backdoor path' between $y_i$ and $t_i$, so that we can measure the casual effect
\begin{equation}
\small
    \delta = \mathbb{E}[\mathbb{E}[y| \bs{Z},\bs{C},t = 1] - \mathbb{E}[y|\bs{Z},\bs{C},t=1] | t=1 ]. \nonumber
\end{equation}
Intuitively, to obtain such $\bs{Z_i}$ and consequently an unbiased treatment effect, one should estimate the text features in a supervised fashion taking into account dependencies between $\bs{W_i}$, $y_i$, $t_i$, and $\bs{C_i}$.
\subsection{\textbf{Estimating Conditional Expectations}} \label{section:CEF}
To estimate the ATT, we need to compute the conditional expectation function (CEF): $\mathbb{E}[\bs y| \bs t,\bs{Z}, \bs{C}]$. Using regression to estimate our conditional expectation function, we can write
\begin{equation}\label{eq_cef_general}
\mathbb{E}[\bs y | \bs t,\bs{Z}, \bs{C}] = f(\bs t,g(\bs{W,\bs y, \bs t, \bs C}; \Theta),\bs{C}; \bs{\Omega}).
\end{equation}
Let $f()$ be the function of our regression equation that we need to define, and $\bs{\Omega}$ be the parameters of it. Section \ref{section_sup_topic_rep} covers text representation function $g()$. For now, 
let us simply assume that we obtain $\bs Z$ in a joint supervised estimation with $f()$. The predominant assumption in causal inference settings in many disciplines is a linear causal effect assumption. We also follow this approach, for the sake of simplicity. However, the requirement for joint supervised estimation of text representations $\bs{Z}$ to be able to predict $\bs y$, $\bs t$ (and if relevant $\bs{C}$) to be considered `causally sufficient' is not constrained to the linear case \cite{veitch2020adapting}. Under the linearity assumption, the CEF of our regression can take the form
\begin{equation}\label{eq_linear_reg_ci}
\bs y = \mathbb{E}[\bs y| \bs t,\bs{Z}, \bs{C}] + \epsilon =  \bs{t} \omega_t + \bs{Z} \bs{\omega_Z}+ \bs{C}\bs{\omega_C} + \epsilon,    
\end{equation}
where $\epsilon \sim N(0,\sigma_\epsilon^2)$ is additive i.i.d. Gaussian. When the CEF is causal, the regression estimates are causal \cite{angrist2008mostly}. In such a case, $\omega_t$ measures the treatment effect. 

\noindent \textbf{Regression Decomposition theorem:} The Frisch-Waugh-Lovell (FWL) theorem \citep{frisch1933partial,lovell1963seasonal}, implies that the supervised learning of text representations $\bs Z$ and regression coefficients $\bs \omega$ cannot be conducted in separate stages, but instead must be learned jointly. 
The FWL theorem states that a regression such as in \eqref{eq_linear_reg_ci} can only be decomposed into separate stages, and still obtain mathematically unaltered coefficient estimates, if for each partial regression, we were able to residualize both outcome and regressors with respect to all other regressors that have been left out. In general, for a regression such as 
$
\bs y = \bs X \bs{\omega} + \bs \epsilon, 
$
we have a projection matrix
$
\bs P = \bs X(\bs X^\intercal \bs X )^{-1}\bs X^\intercal
$
that produces projections $\bs{\widehat{y}}$ when applied to $\bs y$. 
Likewise, we have a `residual maker' matrix $\bs M$ which is $\bs P$'s complement
$
\bs M = \bs I - \bs P.
$
FWL says that if we could estimate
\begin{equation}\label{eq_fwl_alt}
\bs M_{c,z} \bs{y}  = \bs{M}_{c,z} \bs{t}\hat{\omega}_t + \hat{\epsilon},
\end{equation}
the estimates $\hat{\omega}_t$ of treatment effect $\omega_t$ in equations \eqref{eq_linear_reg_ci} and \eqref{eq_fwl_alt} would be mathematically identical (full theorem and proof in Appendix \ref{section:app_ci_text}). Here, $\bs M_{c,z}$ residualizes $\bs t$ from confounders $\bs C$ and $\bs Z$. This is however infeasible, since $\bs Z$ itself must be estimated in a supervised fashion, learning the dependencies towards $\bs y$, $\bs t$ and $\bs C$.
Equation \eqref{eq_linear_reg_ci} must therefore be learned jointly, to infer $\bs Z$ and the CEF in turn.
An approach in several stages in such a setup cannot fully residualize $\bs t$ from all confounders and estimation results would therefore be biased. What is more, if incorrect parameters are learned, out of sample prediction might also be worse. We demonstrate this both on synthetic and semi-synthetic datasets (section \ref{section:synthetic_data} and \ref{section:semi_synthetic_data}).

\subsection{Supervised topic representations}\label{section_sup_topic_rep}
Topic models are a popular choice of text representation in causal inference settings \citep{keith2020text} and in modelling with text as data in social sciences in general \citep{gentzkow2019text}. We focus on this text representation approach for function $g()$ %in equation \eqref{eq_cef_general} 
in our joint modelling strategy. 

\textbf{BTR:} We create BTR, a fully Bayesian supervised topic model that can handle numeric metadata as regression features and labels. Its generative process builds on LDA-based models in the spirit of \citet{blei_mcauliffe2008supervised}.
Given our focus on causal interpretation, we opt for a Gibbs Sampling implementation. This provides statistical guarantees of providing asymptotically exact samples of the target density while (neural) variational inference does not \cite{robertcasella2013MCMC}. \citet{blei2017variational_review} point out that MCMC methods are preferable over variational inference when the aim of the task is to obtain asymptotically precise estimates. 

\textbf{rSCHOLAR:} SCHOLAR \cite{card_2018_SCHOLAR} is a supervised topic model that generalises both sLDA \citep{blei_mcauliffe2008supervised} as it allows for predicting labels, and SAGE \cite{sage_2011} which handles jointly modelling covariates via `factorising' its topic-word distributions ($\beta$) into deviations from the background log-frequency of words and deviations based on covariates. SCHOLAR is solved via neural variational inference \cite{KingmaWellingVAE14, rezende14}. However, it was not primarily designed for causal inference. We extend SCHOLAR with a linear regression layer (rSCHOLAR) to allow direct comparison with BTR. That is, its downstream layer is
$
 \bs y =\bs A \bs \omega,
$
where $\bs A = [\bs t, \bs C, \bs \theta]$ is the design matrix in which $\bs \theta$ represents the estimated document-topic mixtures. $\bs \omega$ represents the regression weight vector. This regression layer is jointly optimized with the main SCHOLAR model via backpropagation using ADAM \cite{kingma_2015_ADAM}, replacing the original downstream cross-entropy loss with mean squared error loss.

Other recent supervised topic models that can handle covariates are for example STM \cite{roberts2016model} and DOLDA \cite{dolda2020}. DOLDA was not designed for regression nor for causal inference setups. Topics models in the spirit of STM incorporate document metadata, but in order to better predict the content of documents rather than to predict an outcome. Many approaches on supervised topic models for regression have been suggested over the years. \cite{blei_mcauliffe2008supervised} optimize their sLDA model with respect to the joint likelihood of the document data and the response variable using VI. MedLDA \cite{zhu2012medlda} optimizes with respect to the maximum margin principle, Spectral-sLDA \cite{wang2014spectral} proposes a spectral decomposition algorithm, and BPsLDA \cite{chen_bpslda_2015} uses backward propagation over a deep neural network. Since BPsLDA reports to outperform sLDA, MedLDA and several other models, we include it in our benchmark list for two-stage models. We include a Gibbs sampled sLDA to have a two-stage model in the benchmark list that is conceptually very similar to BTR in the generative topic modelling part. Unsupervised LDA \cite{blei2003latent, griffiths2004finding} and a neural topic model counterpart GSM \cite{miao2017discovering} are also added for comparison.

%-------------------------------------------------------------
%-------------------------------------------------------------
%-------------------------------------------------------------
\section{Bayesian Topic Regression Model} \label{section:model}
\subsection{Regression Model}\label{section_reg_model}
We take a Bayesian approach and jointly estimate $f()$ and $g()$ to solve equation \eqref{eq_linear_reg_ci}. To simplify notation, encompass numerical features of treatment $\bs t$ and covariates $\bs C$ in data matrix $\bs{X} \in \mathbb{R}^{D \times (1 + \text{dim}_C)}$. All estimated topic features are represented via $\bs{\bar{Z}} \in \mathbb{R}^{D \times K}$, where $K$ is the number of topics. Finally, $\bs y \in \mathbb{R}^{D \times 1}$ is the outcome vector. Define $\bs A=[\bs{\bar{Z}}, \bs X]$ as the overall regression design matrix containing all features (optionally including interaction terms between topics and numerical features).
With our fully Bayesian approach, we aim to better capture feature correlations and model uncertainties. In particular, information from the numerical features (labels, treatment and controls) directly informs the topic assignment process as well as the regression. This counters bias in the treatment effect estimation, following the spirit of `causally sufficient' text representations \cite{veitch2020adapting}. 
Following the previous section, we outline the case for $f()$ being linear.  Our framework could however be extended to non-linear $f()$. 
Assuming Gaussian iid errors $\bs{\epsilon} \sim \mathcal{N}(\bs 0,\sigma^2\bs{I})$, the model's regression equation is then $\bs{y} =  \bs{A}\bs{\omega} + \bs{\epsilon}$, such that
%\small
\begin{equation} \label{eq_regression}
\small
 p(\bs{y}| \bs{A},\bs{\omega},\sigma^2) = \mathcal{N}(\bs{y}|\bs{A}{\bs{\omega}}, \sigma^2\bs{I}). \end{equation}
The likelihood with respect to outcome $\bs{y}$ is then
\begin{equation}\label{eq_likelihood}
\small
p(\bs{y}| \bs{A},\bs{\omega},\sigma^2) = \prod_{d=1}^{D} \mathcal{N}(y_d | \bs{a}_d \bs{\omega}, \sigma^2\bs{I}),
\end{equation}
where $\bs{a}_d$ is the $d$th row of design matrix $\bs{A}$. 
We model our prior beliefs about parameter vector $\bs{\omega}$ by a Gaussian density 
\begin{equation}\label{omega_priors} \small
p(\bs{\omega}) = \mathcal{N}(\bs{\omega}|\bs{m}_0, \bs{S}_0)
\end{equation}
where mean $\bs{m}_0$ and covariance matrix $\bs{S}_0$ are hyperparameters. Following \citet{bishop2006}, we place an Inverse-Gamma prior on the conditional variance estimate $\sigma^2$ with shape and scale
hyperparameters $a_0$ and $b_0$
%\small
\begin{equation}\label{inv_gamma}
\small
    p(\sigma^2) = \mathcal{IG}(\sigma^2 | a_0,b_0).
\end{equation}
Placing priors on all our regression parameters allows us to conduct full Bayesian inference, which not only naturally counteracts parameter over-fitting but also provides us with well-defined posterior distributions over $\bs{\omega}$ and $\sigma^2$ as well as a predictive distribution of our response variable.

Due to the conjugacy of the Normal-Inverse-Gamma prior, the regression parameters' posterior distribution has a known Normal-Inverse-Gamma distribution \cite{stuart1994kendall}
%\small
\begin{equation} \label{eq:omega_posterior}
\small
\begin{split}
p(\bs{\omega},\sigma ^{2} | & \bs{y}, \bs{A} )  \propto 
p(\bs{\omega} | \sigma^{2},\bs{y} ,\bs{A} )p (\sigma ^{2}\mid \bs{y}, \bs{A}) \\
= & \mathcal{N}\left(\bs{\omega} | \bs{m}_{n},\sigma ^{2}\bs{S}_{n}^{-1}\right) {\mathcal{IG}}\left(\sigma^2 | a_{n},b_{n}\right).
\end{split}
\end{equation}
$\bs{m}_{n}$, $\bs{S}_{n}$, $a_{n}$, $b_{n}$  follow standard updating equations for a Bayesian linear regression (Appendix \ref{section:app_reg_model}).

\subsection{Topic Model}\label{section_topic_model}
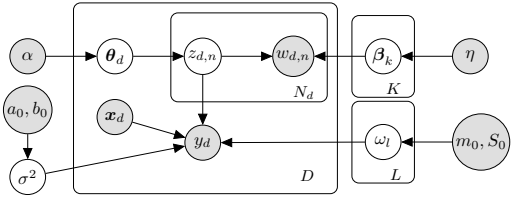
\begin{figure}
	\centering
	\resizebox{6.8cm}{2.55cm}{
	\begin{tikzpicture}
		% nodes
		\node[obs] (alpha) {$\alpha$}; %
		\node[latent,right=of alpha ] (theta) {$\bs{\theta}_d$}; %
		\node[latent,right=of theta ] (z) {$z_{d,n}$}; %
		\node[obs, right=of z] (w) {$w_{d,n}$};%
		\node[latent,right=of w,yshift= 0cm,fill] (beta) {$\bs{\beta}_k$}; %
		\node[obs,right=of beta,xshift=0cm,fill] (eta) {$\eta$}; %
		\node[obs, below=of z, xshift = 0cm, ] (y) {$y_d$};%
		\node[obs, below=of theta, yshift = 0.5cm, ] (x) {$\bs{x}_d$};%
		\node[latent,below=of beta,xshift=0cm] (omega) {$\omega_l$}; %
		\node[obs,right=of omega,xshift=0cm] (m) {$m_0, S_0$}; %
		\node[latent,below=of alpha,yshift=-0.7cm] (sigma) {$\sigma^2$}; %
		\node[obs,above=of sigma,yshift=-0.5cm] (sigma_priors) {$a_0, b_0$}; %
		%\draw (0,1.69) circle(.36cm);
		% plate
		\plate [inner sep=.45cm,yshift=.2cm] {plateD} {(theta)(z)(w)(y)(x)} {$D$}; %
		\plate [inner sep=.25cm,yshift=.2cm] {plateN} {(z)(w)} {$N_d$}; %			
		\plate [inner sep=.25cm,yshift=.2cm] {plateN} {(beta)} {$K$}; %				
		\plate [inner sep=.25cm,yshift=.2cm] {plateL} {(omega)} {$L$}; %				
		% edges
		\edge {alpha} {theta} 
		\edge {theta} {z} 
		\edge {z} {w}  
		\edge {beta} {w}  
		\edge {eta} {beta}  
		\edge {x} {y}
		\edge {z} {y}
		\edge {omega} {y}
		\edge{m} {omega}
		\edge{sigma} {y}
		\edge{sigma_priors} {sigma}
	\end{tikzpicture}
	}
	\caption{Graphical model for BTR.}
	\label{fig:btr_graph}
\end{figure}

The estimated topic features $\bs{\bar{Z}}$, which form part of the design regression matrix $\bs{A}$, are generated from a supervised model that builds on an LDA-based topic structure \cite{blei2003latent}. Figure \ref{fig:btr_graph} provides a graphical representation of BTR and brings together our topic and regression model.

We have $d$ documents in a corpus of size $D$, a vocabulary of $V$ unique words and  $K$ topics. A document has $N_d$ words, so that $w_{d,n}$ denotes the $n$th word in document $d$. The bag-of-words representation of a document is $\bs w_d = [w_{d,1},\dots,w_{d,N_d}]$, so that the entire corpus of documents is described by $\bs{W} = [\bs{w}_1,\dots,\bs{w}_D]$. $z_{d,n}$ is the topic assignment of word $w_{d,n}$, where $\bs{z}_d$ and $\bs{Z}$ mirror $\bs{w}_d$ and $\bs{W}$ in their dimensionality. Similarly, $\bs{\bar{z}}_d$ denotes the estimated average topic assignments of the $K$ topics across words in document $d$, such that $\bs{\bar{Z}}=[\bs{\bar{ z}_1},\dots,\bs{\bar{z}_D}]^\intercal \in \mathbb{R}^{D \times K}$. $\bs{\beta} \in \mathbb{R}^{K \times V}$, describes the $K$ topic distributions over the $V$ dimensional vocabulary.
$\bs{\theta} \in \mathbb{R}^{D \times K}$ describes the $K$ topic mixtures for each of the $D$ documents. 
$\eta \in \mathbb{R}^{V}$ and $\alpha \in \mathbb{R}^K$ are the respective hyperparameters of the prior for $\bs{\beta}$ and $\bs{\theta}$.
The generative process of our BTR model is then:
 \begin{enumerate}[topsep=0.8pt,itemsep=0.2ex,partopsep=0ex,parsep=0.15ex]
	\item $\bs{\omega} \sim \mathcal{N}(\bs{\omega} | \bs{m}_0, \bs{S}_0)$ and $\sigma^2 \sim \mathcal{IG}(\sigma^2 | a_0,b_0)$
	\item \textbf{for} $k = 1,\dots,K$:
	\begin{enumerate} [topsep=0pt,itemsep=-1ex,partopsep=0ex,parsep=1ex]
	    \item $\bs{\beta}_k \sim \text{Dir}(\eta)$
	\end{enumerate}
	\item \textbf{for} {$d = 1,\dots,D$}:
	\begin{enumerate} [topsep=0pt,itemsep=-1ex,partopsep=0ex,parsep=1ex]
	    \item $\bs{\theta}_d \sim \text{Dir}(\alpha)$
	    \item \textbf{for} {$n = 1,\dots,N_d$}:
	    \begin{enumerate}[topsep=-3pt,itemsep=-0.0ex,partopsep=0ex,parsep=0ex]
	        \item topic assignment $z_{d,n} \sim Mult(\bs{\theta}_d)$
	        \item term $w_{d,n} \sim Mult(\bs{\beta}_{z_{d,n}})$
	    \end{enumerate} 
	\end{enumerate}
	\item $\bs{y} \sim \mathcal{N}\left(\bs{A}\bs{\omega}, \sigma^2\bs{I} \right)$.
\end{enumerate}
Straightforward extensions also allow multiple documents per observation or observations without documents, as is described in Appendix \ref{section_obs_wo_docs}.

%-------------------------------------------------------------
\section{Estimation}
\subsection{Posterior Inference}
The objective is to identify the latent topic structure and regression parameters that are most probable to have generated the observed data. We obtain the joint distribution for our graphical model through the product of all nodes conditioned only on their parents, which for our model is
\begin{equation}\label{joint}
\small
\begin{split}
p(\bs{\theta} &, \bs{\beta}, \bs{Z}, \bs{W}, \bs{y}, \bs{\omega}, \sigma^2| \bs{X}, \alpha, \eta, \bs{m}_0, \bs{S}_0, a_0, b_0) = \\
\prod_{d=1}^D & p(\bs{\theta}_d | \alpha) \prod_{k=1}^K p (\bs{\beta}_k | \eta) \prod_{d=1}^D \prod_{n=1}^{N_d} p(z_{d,n} | \bs{\theta}_d)
p(w_{d,n} |z_{d,n}, \bs{\beta})  \\
\prod_{d=1}^D &p(y_d | \bs{x}_d, \bs{z}_d, \bs{\omega}, \sigma^2) \prod_{l=1}^{L} p(\bs{\omega}_{l}|\bs{m}_0, \bs{S}_0) p(\sigma^2 | a_0, b_0).
\end{split}
\end{equation}
The inference task is thus to compute the posterior distribution of the latent variables ($\bs{Z}$, $\bs{\theta}$, $\bs{\beta}$, $\bs{\omega}$, and $\sigma^2$) given the observed data ($\bs{y}$, $\bs{X}$ and $\bs{W}$) and the priors governed by hyperparameters ($\alpha, \eta$, $\bs{m}_0,\bs{S}_0,a_0,b_0$). We will omit hyperparameters for sake of clarity unless explicitly needed for computational steps. The posterior distribution is then
\begin{equation}\label{posterior} \small
p(\bs{\theta}, \bs{\beta}, \bs{Z}, \bs{\omega}, \sigma^2 | \bs{W}, \bs{y}, \bs{X}) = \frac{p(\bs{\theta}, \bs{\beta}, \bs{Z}, \bs{W}, \bs{X}, \bs{y}, \bs{\omega}, \sigma^2 )}{p(\bs{W},\bs{X},\bs{y})}.
\end{equation}
In practice, computing the denominator in equation \eqref{posterior}, i.e. the evidence, is intractable due to the sheer number of possible latent variable configurations. 
We use a Gibbs EM algorithm \cite{levine2001implementations} set out below, to approximate the posterior. Collapsing out the latent variables $\bs{\theta}$ and $\bs{\beta}$ \cite{griffiths2004finding}, we only need to identify the sampling distributions for topic assignments $\bs{Z}$ and regression parameters $\bs{\omega}$ and $\sigma^2$, conditional on their Markov blankets 
\begin{equation}
\small
\begin{split}
& p(\bs{Z}, \bs{\omega}, \sigma^2 | \bs{W}, \bs{X}, \bs{y}) = \\
& p(\bs{Z} | \bs{W}, \bs{X}, \bs{y}, \bs{\omega}, \sigma^2)p(\bs{\omega}, \sigma^2 | \bs{Z}, \bs{X}, \bs{y}).
\end{split}
\end{equation}
Once topic assignments $\bs{Z}$ are estimated, it is straightforward to recover $\bs{\beta}$ and $\bs{\theta}$. The expected topic assignments are estimated by Gibbs sampling in the E-step, and the regression parameters are estimated in the M-step.
\subsection{E-Step: Estimate Topic Parameters} \label{section:E_Step}
In order to sample from the conditional posterior for each $z_{d,n}$ we need to identify the probability of a given word $w_{d,n}$ being assigned to a given topic $k$, conditional on the assignments of all other words (as well as the model's other latent variables and the observed data)
%\small
\begin{equation}
\small
p(z_{d,n} = k | \bs{Z}_{-(d,n)}, \bs{W}, \bs{X}, \bs{y}, \bs{\omega}, \sigma^2),
\end{equation}
where $\bs{Z}_{-(d,n)}$ are the topic assignments of all words apart from $w_{d,n}$. This section defines this distribution, with derivations in Appendix \ref{section:app_topic_model}. By conditional independence properties of the graphical model, we can split this joint posterior into
\begin{equation}
\small \label{sLDA_cov_posterior_form}
p(\bs{Z}|\bs{W}, \bs{X}, \bs{y}, \bs{\omega}, \sigma^2) \propto p(\bs{Z}|\bs{W}) p(\bs{y}|\bs{Z}, \bs{X}, \bs{\omega}, \sigma^2).
\end{equation}
Topic assignments within one document are independent from topic assignments in all other documents and the sampling equation for $z_{d,n}$ only depends on it's own response variable $y_d$, hence 
\begin{equation}
\label{eq:z_sampling_form}
\small
\begin{split}
& p(z_{d,n} = k | \bs{Z}_{-(d,n)}, \bs{W}, \bs{X}, \bs{y}, \bs{\omega}, \sigma^2)  \propto \\ 
& p(z_{d,n} = k | \bs{Z}_{-(d,n)}, \bs{W}) p(y_d|z_{d,n}=k, \bs{Z}_{-(d,n)}, \bs{x}_d, \bs{\omega}, \sigma^2).
\end{split}
\end{equation}

The first part of the RHS expression is the sampling distribution of a standard LDA model. Following \citet{griffiths2004finding}, we can express it in terms of count variables $s$ (topic assignments across a document) and $m$ (assignments of unique words across topics over all documents).\footnote{For example, $s_{d,k}$ denotes the total number of words in document $d$ assigned to topic $k$ and $s_{d,k,-n}$ the number of words in document $d$ assigned to topic $k$, except for word $n$. Analogously, $m_{k,v}$ measures the total number of times term $v$ is assigned to topic $k$ across all documents and $m_{k,v,-(d,n)}$ measures the same, but excludes word $n$ in document $d$.}

The second part is the predictive distribution for $y_d$. This is a Gaussian distribution depending on the linear combination $\bs{\omega}(\bs{a}_d|z_{d,n} = k)$, where $\bs{a}_d$ includes the topic proportions $\bs{\bar{z}}_d$ and $\bs{x}_d$ variables (and any interaction terms), conditional on $z_{d,n} = k$. We can write this in a convenient form that preserves proportionality with respect to $z_{d,n}$ and depends only on the data and the count variables.

First, we split the $\bs{X}$ features into those that are interacted, $\bs{X}_{1,d}$, and those that are not, $\bs{X}_{2,d}$ such that the generative model for $y_d$ is then
%\
\begin{equation}
\small
y_d \sim \mathcal{N}(\bs{\omega}_z^\intercal \bs{\bar{z}}_{d} + \bs{\omega}_{zx}^\intercal (\bs{x}_{1,d} \otimes \bs{\bar{z}}_d) + \bs{\omega}_x^\intercal \bs{x}_{2,d}, \sigma^2),
\end{equation}
where $\otimes$ is the Kronecker product.
Define $\bs{\tilde{\omega}}_{z,d}$ as a length $K$ vector such that
\begin{equation} 
\small
\tilde{\omega}_{z,d,k} = \omega_{z,k} + \bs{\omega}_{zx,k}^\intercal \bs{x}_{1,d}.
\end{equation}
Noting that $\bs{\tilde{\omega}}_{z,d}^\intercal \bs{\bar{z}}_d = \frac{\bs{\tilde{\omega}}_{z,d}^\intercal}{N_d}(\bs{s}_{d,-n} + s_{d,n})$, gives us the sampling distribution for $z_{d,n}$ stated in equation \eqref{eq:z_sampling_form}: a multinomial distribution parameterised by 
\begin{align} \scriptsize
& p(z_{d,n} = k | z_{-(d,n)}, W,X,y, \alpha, \eta, \omega, \sigma^2) \propto \nonumber \\ 
& (s_{d,k, -n } + \alpha) \times
\frac{m_{k,v, -(d,n)} +\eta}{\sum_v m_{k,v, -(d,n)} +V \eta} \nonumber \\
& \exp \Bigg\{ \frac{1}{2 \sigma^2} \bigg( \frac{2 \tilde{\omega}_{z,d,k}}{N_d} \left(y_d - \nonumber \bs{\omega}_x^\intercal \bs{x}_{2,d} - \frac{\bs{\tilde{\omega}}_{z,d}^\intercal}{N_d}\bs{s}_{d,-n}\right) \nonumber \\
& - \left(\frac{\tilde{\omega}_{z,d,k}}{N_d} \right) ^2 \bigg) \Bigg\}.
\end{align}
This defines the probability for each $k$ that $z_{d,n}$ is assigned to that topic $k$. These $K$ probabilities define the multinomial distribution from which $z_{d,n}$ is drawn.

\subsection{M-Step: Estimate Regression Parameters}
To estimate the regression parameters, we hold the design matrix $\bs{A} = [\bs{\bar{Z}},\bs{X}]$ fixed. Given the Normal-Inverse-Gamma prior, this is a standard Bayesian linear regression problem and the posterior distribution for which is given in equation \eqref{eq:omega_posterior} above. To prevent overfitting to the training sample there is the option to randomly split the training set into separate sub-samples for the E- and M-steps, following a Cross-Validation EM approach \cite{shinozaki2007cross}. We use the prediction mean squared error from the M-step sample to assess convergence across EM iterations.

\subsection{Implementation}
We provide an efficient \textit{Julia} implementation for BTR and a \textit{Python} implementation for rSCHOLAR on Github to allow for reproducibility of the results in the following experiment sections.\footnote{BTR: \href{https://github.com/julianashwin/BTR.jl}{github.com/julianashwin/BTR.jl} \\ rSCHOLAR:  \href{https://github.com/MaximilianAhrens/scholar4regression}{github.com/MaximilianAhrens/scholar4regression}}

%-------------------------------------------------------
%-------------------------------------------------------------
%-------------------------------------------------------------
\section{Experiment: Synthetic Data}\label{section:synthetic_data}
\subsection{Synthetic Data Generation}

To illustrate the benefits of our BTR approach, we generate a synthetic dataset of documents which have explanatory power over a response variable, along with an additional numerical covariate that is correlated with both documents and response.

\begin{figure*}[t!]
\centering % <-- added
\begin{subfigure}{0.25\textwidth}
  \includegraphics[width=\linewidth]{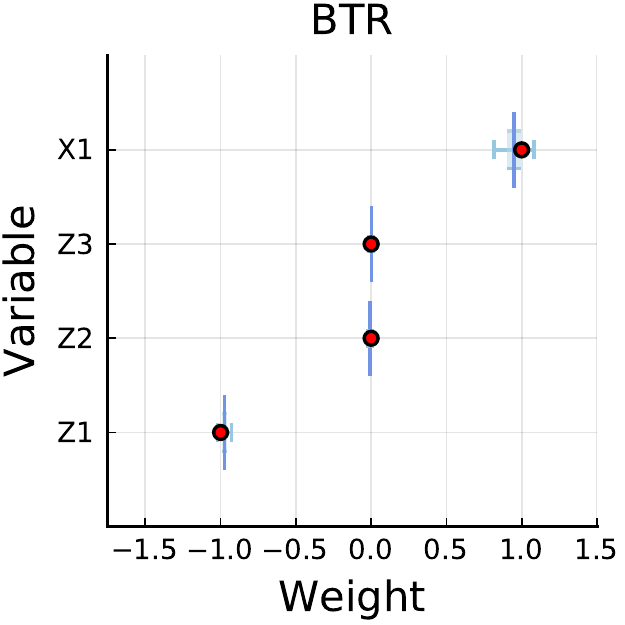}
  \label{fig:synth_btr}
\end{subfigure}\hfil % <-- added
\begin{subfigure}{0.25\textwidth}
  \includegraphics[width=\linewidth]{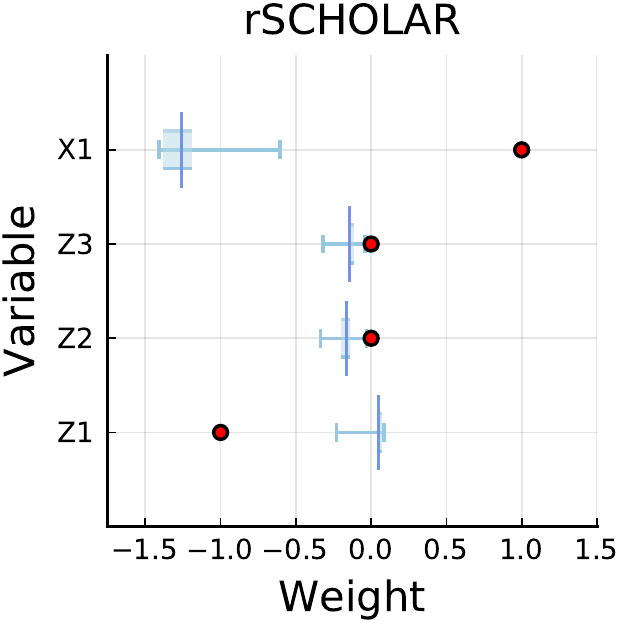}
  \label{fig:synth_scholar}
\end{subfigure}\hfil % <-- added
\begin{subfigure}{0.25\textwidth}
  \includegraphics[width=\linewidth]{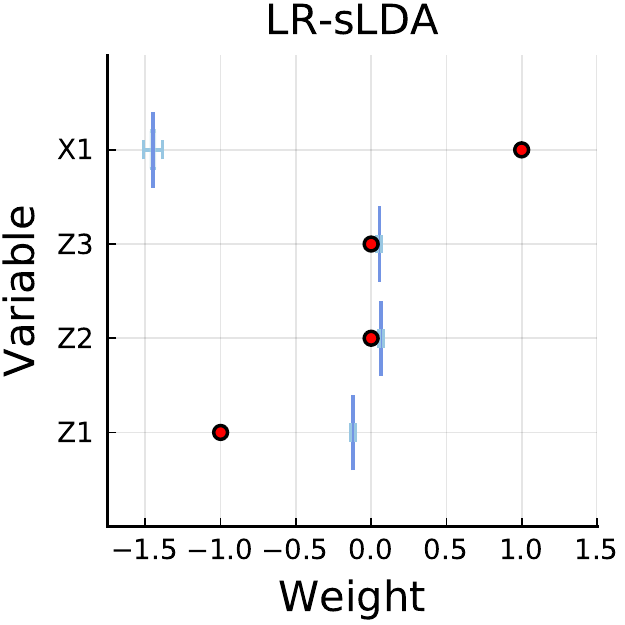}
  \label{fig:synth_lr_slda}
\end{subfigure}
\medskip
\begin{subfigure}{0.25\textwidth}
  \includegraphics[width=\linewidth]{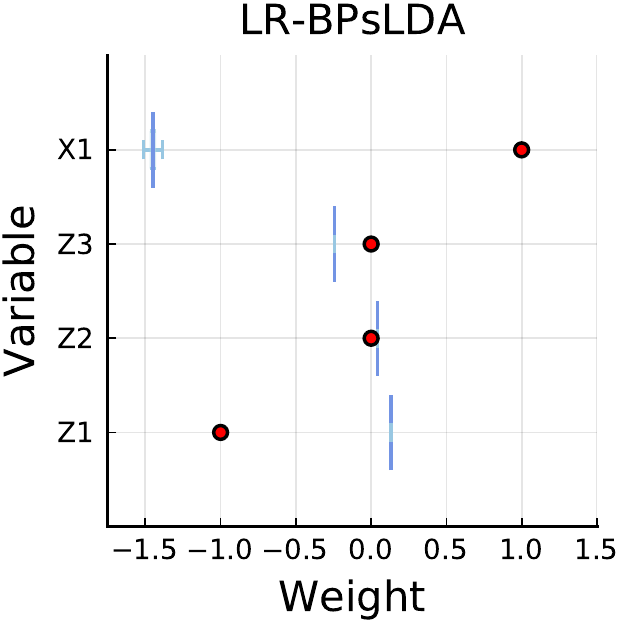}
  \label{fig:synth_lr_bpslda}
\end{subfigure}\hfil % <-- added
\begin{subfigure}{0.25\textwidth}
  \includegraphics[width=\linewidth]{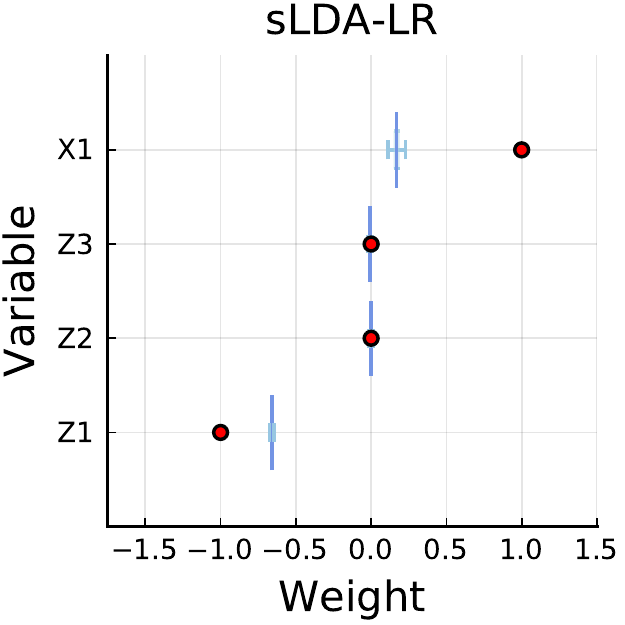}
  \label{fig:synth_slda_lr}
\end{subfigure}\hfil % <-- added
\begin{subfigure}{0.25\textwidth}
  \includegraphics[width=\linewidth]{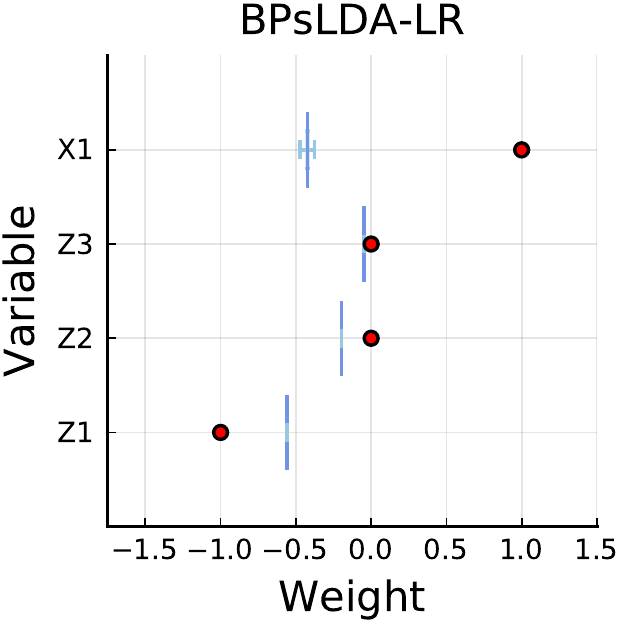}
  \label{fig:synth_bpslda_lr}
\end{subfigure}
\caption{Comparing recovery of true regression weights across different topic models. For each panel, the true regression weights are shown as red points and the estimated $95\%$ posterior credible (or bootstrap, depending on model) interval in blue. Only BTR contains the true weights within the estimated intervals.}
\label{fig:synth_data_model_comparion}
\end{figure*}
We generate $10,000$ documents of $50$ words each, following an LDA generative process, with each document having a distribution over three topics, defined over a vocabulary of 9 unique terms. A numerical feature, $\bs{x}=[x_1,...,x_D]^\intercal $, is generated by calculating the document-level frequency of the first word in the vocabulary. As the first topic places a greater weight on the first three terms in the vocabulary, $\bs{x}$ is positively correlated with $\bs{\bar{z}}_{1}$. The response variable $\bs{y}=[y_1,...,y_D]$ is generated through a linear combination of the numerical feature $\bs{x}$ and the average topic assignments $\bs{\bar{Z}}=\{\bs{\bar{z}}_{1},\bs{\bar{z}}_{2},\bs{\bar{z}}_{3}\}$,
\begin{equation} \label{eq:y_gen}
\small
    \bs{y} =  - \bs{\bar{z}}_{1} + \bs{x} + \bs{\epsilon}.
\end{equation}
where $\bs{\epsilon}$ is an iid Gaussian white noise term. The regression model to recover the ground truth is then
\begin{equation} \label{eq:y_gen}
\small
    \bs y =  \omega_1 \bs{\bar{z}}_{1} + \omega_2 \bs{\bar{z}}_{2} + \omega_3 \bs{\bar{z}}_{3} + \omega_4 \bs x_{d} + \bs{\epsilon}.
\end{equation}
The \textit{true} regression weights are thus $\bs{\omega}^* = [-1, 0, 0,1]$. In accordance with the FWL theorem, we cannot recover the true coefficients with a two-stage estimation process.

\subsection{Synthetic Data Results}\label{section:synth_results}
We compare the ground truth of the synthetic data generating process against: (1) \textbf{BTR:} our Bayesian model, estimated via Gibbs sampling. (2) \textbf{rSCHOLAR}: the regression extension of SCHOLAR, estimated via neural VI. (3) \textbf{LR-sLDA:} first linearly regress $\bs{y}$ on $\bs{x}$, then use the residual of that regression as the response in an sLDA model, estimated via Gibbs sampling. (4) \textbf{sLDA-LR:} First sLDA, then linear regression. (5) \textbf{BPsLDA-LR} and (6)\textbf{ LR-BPsLDA:} replace sLDA with BPsLDA, which is sLDA estimated via the backpropagation approach of \citet{chen_bpslda_2015}.

Figure \ref{fig:synth_data_model_comparion} shows the true and estimated regression weights for each of the six models. LR-sLDA and sLDA-LR estimate inaccurate regression weights for both the text and numerical features, as do the BPsLDA variants. Similarly, rSCHOLAR fails to recover the ground truth. However, BTR estimates tight posterior distributions around to the true parameter values. The positive correlation between $\bs{z}_{1}$ and $\bs{x}$ makes a joint estimation approach crucial for recovering the true parameters. Standard supervised topic models estimate the regression parameters for the numerical features separately from the topic proportions and their associated regression parameters, violating the FWL theorem as outlined in section \ref{section:CEF}. A key difference between rSCHOLAR and BTR lies in their posterior estimation techniques (neural VI vs Gibbs). rSCHOLAR's approach seems to have a similarly detrimental effect as the two-stage approaches. We suspect further research into (neural) VI assumptions and their effect on causal inference with text could be fruitful. 
%-------------------------------------------------------------
%-------------------------------------------------------------
%-------------------------------------------------------------
\section{Experiment: Semi-Synthetic Data} \label{section:semi_synthetic_data}
\subsection{Semi-Synthetic Data Generation}

We further benchmark the models' abilities to recover the ground truth on two semi-synthetic datasets. We still have access to the ground truth (GT) as we either synthetically create or directly observe the correlations between treatment, confounders and outcome. However, the text and some numeric metadata that we use is empirical. We use customer review data from \textbf{(i) Booking.com}\footnote{Available at \href{https://www.kaggle.com/jiashenliu}{kaggle.com/jiashenliu}} and \textbf{(ii) Yelp}\footnote{Available at \href{https://www.yelp.com/dataset}{yelp.com/dataset}, Toronto subsample}, and analyse two different `mock' research questions. For both datasets, we randomly sample $50,000$ observations and select $75\%$ in Yelp, $80\%$ in Booking for training.\footnote{Appendix \ref{section_app_real_world_data} for full data summary statistics.  Data samples used for experiments available via: \href{https://github.com/MaximilianAhrens/data}{github.com/MaximilianAhrens/data}}
\\~\\
\noindent \textbf{Booking:}
\textit{Do people give more critical ratings ($y_i$) to hotels that have high historic ratings ($av\_score_i$), once controlling for review texts?}
\begin{equation}
\small
GT_B:\quad  y_i =  - \text{hotel\_av}_i + 5\text{prop\_pos}_i
\end{equation}
where $\text{prop\_pos}_i$ is the proportion of positive words in a review. The textual effect is estimated via topic modelling in our experiment. The treatment in question is the average historic customer rating, being modelled as continuous.
\\~\\
\noindent \textbf{Yelp:}
\textit{Do people from the US ($\text{US}_i$=1) give different Yelp ratings ($y_i$) than customers from Canada ($\text{US}_i$=0), controlling for average restaurant review ($\text{stars\_av\_b}_i$) and the review text?}
\begin{equation}
\small
GT_Y: y_i = - \text{US}_i + \text{stars\_av\_b}_i + sent_i.
\end{equation}
To create the binary treatment variable $\text{US}_i$, we compute each review's sentiment score ($sent_i$) using the Harvard Inquirer. This treatment effect is correlated with the text as
\begin{equation}
\small
\Pr(US_i = 1) = \frac{\exp(\gamma_1 sent_i)}{1 + \exp(\gamma_1 sent_i)},
\end{equation}
where $\gamma_1$ controls the correlation between text and treatment.\footnote{When $\gamma_1 =1 $, correlation between $US_i$ and $sent_i$ is $0.23$. For $\gamma_1 =0.5$ it is $0.39$.}

\begin{figure*}
    \centering
    \subfloat{
    \includegraphics[trim=0cm 0 0 0, clip, height = 0.7 \columnwidth]{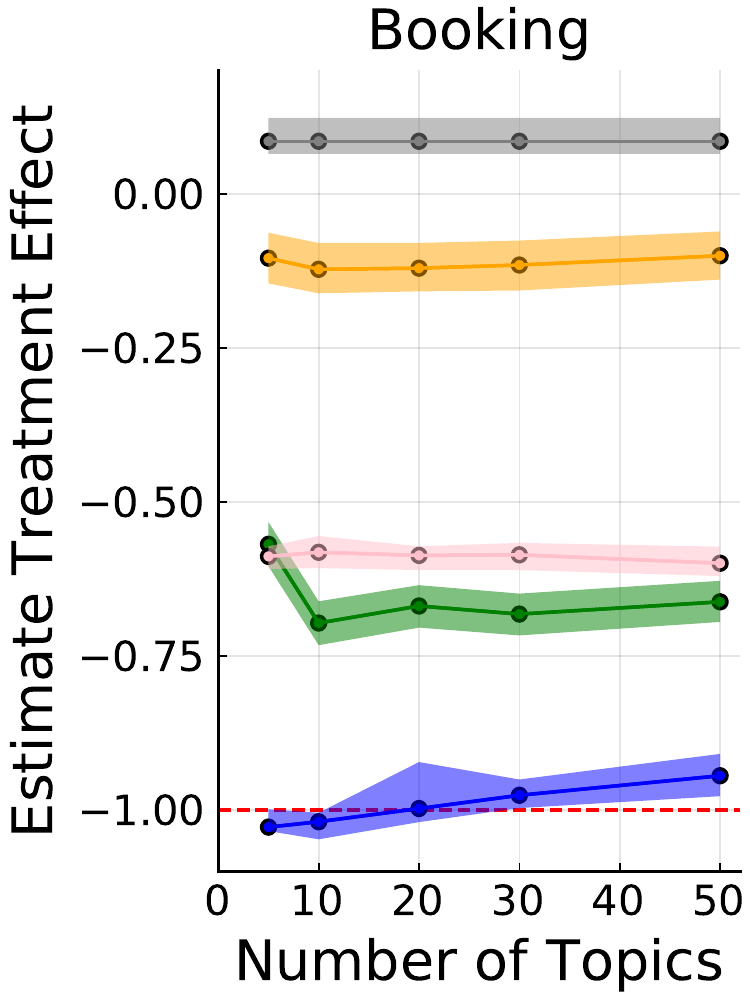}}
    \subfloat{
    \includegraphics[trim=2.1cm 0 0 0, clip, height = 0.7 \columnwidth]{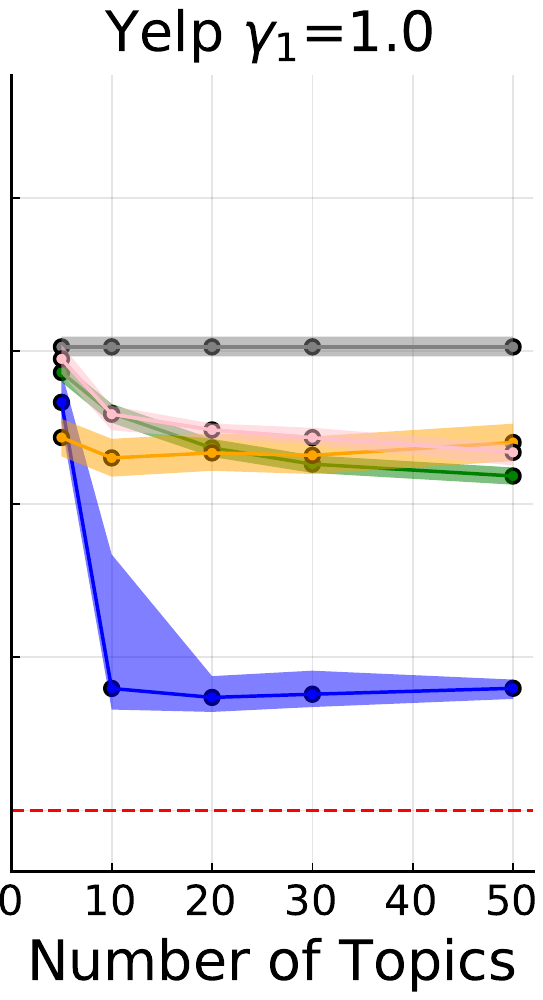}}
    \subfloat{
    \includegraphics[trim=2.1cm 0 0 0, clip, height = 0.7 \columnwidth]{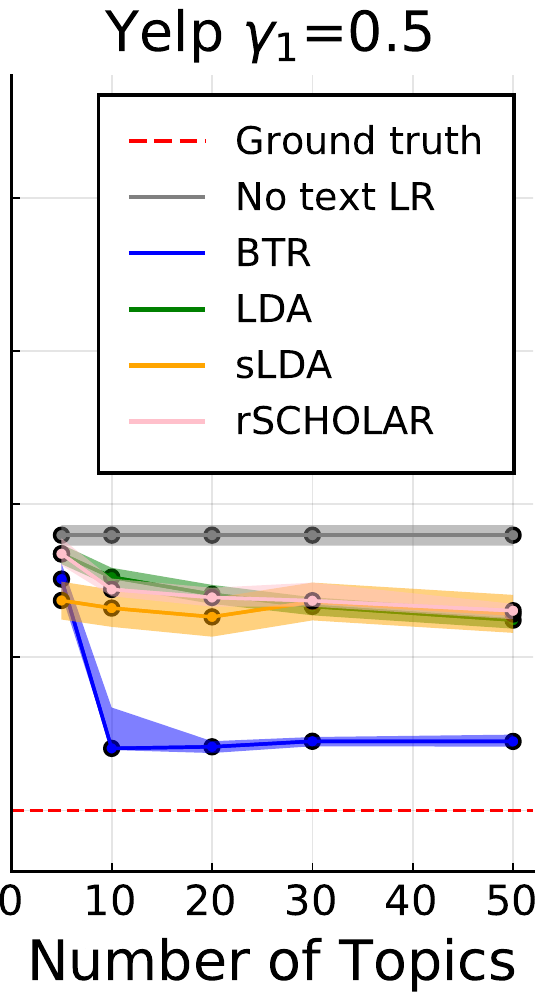}}
    \caption{Estimated TE semi-synthetic Booking (left panel), Yelp (middle and right panel). Intervals are either 95\% credible interval of posterior distribution, or based on 20 run bootstrap, depending on model.}
    \label{fig:groundtruth_semi_synth}
\end{figure*}
\subsection{Semi-Synthetic Data Results}

On both semi-synthetic datasets and across all benchmarked models, BTR estimates the regression weights that are the closest to the ground truth. This consistently holds true across all tested numbers of topics $K$ (see Figure \ref{fig:groundtruth_semi_synth}). For Yelp, we also vary the correlation strength between treatment and confounder. The middle panel in Figure  \ref{fig:groundtruth_semi_synth} shows the estimation results with a very high correlation between confounder and treatment ($\gamma_1 = 1$). The RHS panel shows the results when this correlation is lower ($\gamma_1 = 0.5$). As expected, a higher correlation between confounder and treatment increases the bias as outlined in the section \ref{section:CEF}. If the correlation between confounder and treatment is zero, a two-stage estimation approach no longer violates FWL and all models manage to estimate the ground truth (see Appendix \ref{section:app_semisynth}). Since the topic modelling approach is an approximation to capture the true effect of the text and its correlation with the metadata - and since this approximation is not perfect - some bias may remain. Overall, BTR gets substantially closer to the ground truth than any other model.

%-------------------------------------------------------------
%-------------------------------------------------------------
%-------------------------------------------------------------
\section{Experiment: Real-World Data}\label{section:emp_experiments}

\begin{table}[t]
\centering
\resizebox{\linewidth}{!}{
    \begin{tabular}{@{\extracolsep{0pt}} lcc}
    \toprule
    \textit{Dataset} & \multicolumn{2}{c}{Booking}\\
    \midrule
    \textit{K}  & 50 & 100\\
    \midrule
    {$pR^2$ (higher is better)} \\
    \midrule
    OLS & \multicolumn{2}{c}{0.315} \\
    aRNN  & \multicolumn{2}{c}{0.479 (0.007)}  \\
    LR+ TAM & 0.479 (0.014) & 0.487 (0.014)\\
    LDA+LR &  0.426 (0.003) & 0.437 (0.002) \\
    GSM+LR &  0.386	(0.004) & 0.395 (0.005)\\
    LR+sLDA & 0.432 (0.002) & 0.438 (0.004)\\
    LR+BPsLDA &  0.419 (0.009) & 0.455 (0.001)\\    
    LR+rSCHOLAR & 0.469 (0.002) & 0.465 (0.002)\\
    \textbf{rSCHOLAR} & \textbf{0.494} (0.004)& \textbf{0.489} (0.003)\\
    \textbf{BTR} & 0.454 (0.003)&  0.460 (0.002)\\
    \bottomrule
    \\[-1.8ex] 
    {Perplexity (lower is better)} \\
    \midrule
    LR+TAM & 521 (2) & 522 (2)\\
    LDA+LR &  454 (1) & 432 (1)\\
    GSM+LR & \textbf{369} (8) & \textbf{348} (5) \\
    LR+sLDA & 436 (2) & 411 (1) \\
    LR+rSCHOLAR & 441 (20) & 458 (11)\\
    \textbf{rSCHOLAR} & 466 (19) & 464 (9) \\
    \textbf{BTR} &  437 (1) & 412 (1) \\
    \bottomrule
    \end{tabular}}
    \caption{Booking: mean $pR^2$ and perplexity, standard deviation in brackets. 20 model runs. Best model \textbf{bold}.}
    \label{tab:results_empirical}
\end{table}

\begin{table}[t]
\centering
\resizebox{\linewidth}{!}{
    \begin{tabular}{@{\extracolsep{0pt}} lcc}
    \toprule
    \textit{Dataset} & \multicolumn{2}{c}{Yelp} \\
    \midrule
    \textit{K} & 50 & 100  \\
    \midrule
    {$pR^2$ (higher is better)} \\
    \midrule
    OLS  & \multicolumn{2}{c}{0.451} \\
    aRNN & \multicolumn{2}{c}{0.582 (0.008)}  \\
    LR+ TAM  & 0.585 (0.012) & 0.587 (0.008) \\
    LDA+LR  & 0.586 (0.006) & 0.606 (0.007) \\
    GSM+LR  & 0.495 (0.004) & 0.517 (0.007) \\
    LR+sLDA & 0.571 (0.002) & 0.574 (0.001) \\
    LR+BPsLDA & 0.603 (0.002) & 0.609 (0.001) \\    
    LR+rSCHOLAR & 0.550 (0.034) & 0.557 (0.027) \\
    \textbf{rSCHOLAR} & 0.571 (0.01) & 0.581 (0.009) \\
    \textbf{BTR} & \textbf{0.630} (0.001) & \textbf{0.633} (0.001)\\
    \bottomrule
    \\[-1.8ex] 
    {Perplexity (lower is better)} \\
    \midrule
    LR+TAM & 1661 (7) & 1655 (7) \\
    LDA+LR & 1306 (4) &  1196 (2)\\
    GSM+LR & 1431 (34) & 1387 (14)   \\
    LR+sLDA & 1294 (5) & 1174 (3)  \\
    LR+rSCHOLAR & 1515 (34) & 1516 (30) \\
    \textbf{rSCHOLAR} & 1491 (9) & 1490 (9) \\
    \textbf{BTR} & \textbf{1291 (5)} &  \textbf{1165} (3)  \\
    \bottomrule
    \end{tabular}}
    \caption{Yelp: mean $pR^2$ and perplexity, standard deviation in brackets. 20 model runs. Best model \textbf{bold}.}
    \label{tab:results_empirical}
\end{table}

The joint supervised estimation approach using text and non-text features, not only counteracts bias in causal settings. It also improves prediction performance. We use the real-world datasets of Booking and Yelp for our benchmarking. For both datasets, we predict customer ratings (response) for a business or hotel given customer reviews (text features) and business and customer metadata (numerical features).\footnote{full specifications for each case are given in Appendix \ref{section_app_real_world_data}} 

\subsection{Benchmarks}
We add the following models to the benchmark list from the previous section:\footnote{We also tested sLDA+LR and a pure sLDA, which performed consistently worse, see Appendix \ref{app_empirical_data_results}} \textbf{LDA+LR} \cite{griffiths2004finding} and \textbf{GSM+LR} \cite{miao2017discovering} unsupervised Gibbs sampling and neural VI based topic models. \textbf{LR+rSCHOLAR}: the two-step equivalent for rSCHOLAR, estimating covariate regression weights in a separate step from the supervised topic model.

An alternative to topic based models are word-embedding based neural networks. We use (7) \textbf{LR+aRNN}: a bidirectional RNN with attention \cite{bahdanauCB15}. Since the model does not allow for non-text features, we use the regression residuals of the linear regression as the target. And (8) \textbf{LR+TAM}: a bidirectional RNN using global topic vector to enhance its attention heads \cite{wang2020neural} - same target as in LR+aRNN. 
\footnote{\citet{wang2020neural} use $100$-dimensional word embeddings in their default setup for TAM and pre-train those on the dataset. We follow this approach. RNN and TAM results were very robust to changes in the hidden layer size in these setups, we use a layer size of $64$. Full details of all model parametrisations are provided in Appendix \ref{app:model_params}.} 

\subsection{Prediction and Perplexity Results}
We evaluated all topic models on a range from 10 to 100 topics, with results for 50 and 100 in Table \ref{tab:results_empirical}.\footnote{ Hyperparameters of displayed results: $\alpha=0.5$, $\eta=0.01$} Hyperparameters of benchmark models that have no direct equivalent in our model were set as suggested in the pertaining papers. We find that our results are robust across a wide range of hyperparameters (extensive robustness checks in Appendix \ref{section:app_realworld_experiments}).

We assess the models' predictive performance based on predictive $R^2$ ($pR^2= 1 - \frac{\text{MSE}}{var(y)}$). The upper part of Table \ref{tab:results_empirical} shows that BTR achieves the best $pR^2$ in the Yelp dataset and and very competitive results in the Booking dataset, where our rSCHOLAR extension outperforms all other models. Even the non-linear neural network models aRNN and TAM cannot achieve better results. Importantly, rSCHOLAR and BTR perform substantially better than their counterparts that do not jointly estimate the influence of covariates (LR+rSCHOLAR and LR+sLDA).

To assess document modelling performance, we report the test set perplexity score for all models that allow this (Table \ref{tab:results_empirical}, bottom panel) . Perplexity is defined as $\exp\left\{ -\frac{\sum^{D}_{d=1} \log p(\bs{w}_d|\bs{\theta},\bs{\beta})}{\sum_{d=1}^{D}N_d}\right\}$. The joint approach of both rSCHOLAR and BTR does not come at the cost of increased perplexity. If anything, the supervised learning approach using labels and covariates even improves document modelling performance when compared against its unsupervised counterpart (BTR vs LDA). 

Assessing the interpretability of topic models is ultimately a subjective exercise. In Appendix \ref{section:app_topics} we show topics associated with the most positive and negative regression weights, for each dataset. Overall, the identified topics and the sign of the associated weights seem interpretable and intuitive.

%-------------------------------------------------------------
%-------------------------------------------------------------
%-------------------------------------------------------------
\section{Conclusions}
In this paper, we introduced BTR, a Bayesian topic regression framework that incorporates both numerical and text data for modelling a response variable, jointly estimating all model parameters. Motivated by the FWL theorem, this approach is designed to avoid potential bias in the regression weights, and can provide a sound regression framework for statistical and causal inference when one needs to control for both numerical and text based confounders in observational data. We demonstrate that our model recovers the ground truth with lower bias than any other benchmark model on synthetic and semi-synthetic datasets.
Experiments on real-world data show that a joint and supervised learning strategy also yields superior prediction performance compared to `two-stage' strategies, even competing with deep neural networks.

\section*{Acknowledgements}
Maximilian Ahrens and Julian Ashwin were supported by the Economic and Social Research Council of the UK. Maximilian Ahrens, Jan-Peter Calliess and Vu Nguyen are furthermore grateful for support from the Oxford-Man-Institute.

\bibliography{btr_arxiv}
\bibliographystyle{acl_natbib}

\appendix
\onecolumn

\section{Causal Inference with Text}\label{section:app_ci_text}
For $D$ observations, we have outcome $\bs y \in \mathbb{R}^{D\times1}$, treatment $\bs t \in \mathbb{R}^{D\times1}$, text data $\bs W \in \mathbb{R}^{D\times V}$ (where $V$ is the vocabulary size) and numerical confounders $ \bs C \in \mathbb{R}^{D\times P}$ (where $P$ is the number of numerical confounders).

As established in the main part of the paper, in order to estimate the ATT, we need to compute the conditional expectation function (CEF) $\mathbb{E}[\bs y| \bs t,\bs{Z}]$ or if we have additional numerical confounders $\mathbb{E}[\bs y| \bs t,\bs{Z}, \bs{C}]$. Using regression to estimate our conditional expectation function, we can write
\begin{equation}
\mathbb{E}[\bs y | \bs t,\bs{Z}, \bs{C}] = f(\bs t,\bs{Z},\bs{C}; \bs{\Omega}).
\end{equation}
Let $f()$ be the function of our regression equation that we need to define, and $\bs{\Omega}$ be the parameters of it. The predominant assumption in causal inference settings in many disciplines is a linear causal effect assumption. We follow this approach, also for the sake of simplicity. However, the requirement for joint supervised estimation of text representations $\bs{Z}$ to be able to predict $\bs y$,$\bs t$ (and if relevant $\bs{C}$) to be considered `causally sufficient' is not constrained to the linear case \cite{veitch2020adapting}. Under the linearity assumption, the CEF of our regression can take the form
\begin{equation}\label{app_eq_linear_reg_ci}
\bs y = \mathbb{E}[\bs y| \bs t,\bs{Z}, \bs{C}] + \epsilon =  \bs{t} \omega_t + \bs{Z} \bs{\omega_Z}+ \bs{C}\bs{\omega_C} + \epsilon,    
\end{equation}
where $\epsilon \sim N(0,\sigma_\epsilon^2)$ is additive iid Gaussian noise, ie. $\mathbb{E}[\epsilon|  \bs{t},\bs{Z}, \bs{C}]=0$ (see for example \citet{angrist2008mostly}, chapter 3). Thus, $\sigma_\epsilon$ represents the conditional variance $Var(\bs y| \bs t,\bs{Z},\bs{C})$. 
The regression approximates the CEF. Hence, when the CEF is causal, the regression estimates are causal \cite{angrist2008mostly}. In such a case, $\omega_t$ measures the treatment effect. Assuming that $\bs Z$ and $\bs C$ block all `backdoor' paths, the CEF would allow us to conduct causal inference of the ATT of $\bs t$ on $\bs y$ \cite{pearl_2009causality}.

We now shall revisit under which conditions, a decomposition of equation \eqref{app_eq_linear_reg_ci} into several separate estimation steps is permitted as described in the Frisch-Waugh-Lovell (or regression decomposition) theorem \cite{lovell2008FWL}, so that the regression estimates for $\omega_t$ remain unchanged and hence can still be considered as causal.

\subsection{Regression Decomposition Theorem}
The regression decomposition theorem or Frisch-Waugh-Lovell (FWL) theorem \cite{frisch1933partial, lovell1963seasonal} states that the coefficients of a linear regression as stated in equation \eqref{app_eq_linear_reg_ci} are equivalent to the coefficients of partial regressions in which the residualized outcome is regressed on the residualized regressors - this residualization is in terms of all regressors that are not part of this partial regression.

For a moment, let us assume there are no confounding latent (that is to be estimated) text features $\bs Z$.  Our observational data only consist of outcome $\bs y$, our treatment variable $\bs t$ and other observed confounding variables $\bs C$,
\begin{equation}\label{app_eq_linear_reg_ci_no_Z}
\bs y  =  \bs{t} \omega_t + \bs{C}\bs{\omega_C} + \epsilon. 
\end{equation}
The FWL theorem states that we would obtain mathematically identical regression coefficients $\omega_t$ and $\bs \omega_C$ is we decomposed this regression and estimated each part separately, each time residualizing (ie. orthogonalizing) outcomes and regressors on all other regressors. 
\\~\\
More generally, for a linear regression define
$$
\bs y = \bs X \bs \beta + \bs \epsilon
$$
with $\bs y \in \mathbb{R}^{D\times 1}$, $\bs \beta \in \mathbb{R}^{K\times 1}$, $\bs X \in \mathbb{R}^{D\times M}$, which we could arbitrarily partition into $\bs X_1 \in \mathbb{R}^{D \times K}$ and $\bs X_2 \in \mathbb{R}^{D \times J}$ so we could also write 
$$
\bs y = \bs X_1 \bs \beta_1 + \bs X_2 \bs \beta_2 +\bs \epsilon,
$$
define projection (or prediction) matrix $\bs P$ such that
\begin{equation}
\bs P = \bs X(\bs X^\intercal \bs X )^{-1}\bs X^\intercal.
\end{equation}
$\bs P$ produces predictions $\bs{\widehat{y}}$ when applied to outcome vector $\bs y$,
\begin{equation}
\bs{\widehat{y}} = \bs X \bs{\widehat{\beta}} = \bs X(\bs X^\intercal \bs X )^{-1}\bs X^\intercal \bs y = \bs P \bs y.
\end{equation}
Also define the complement of $\bs P$, the residual maker matrix $\bs M$
\begin{equation}\label{eq_residual_maker}
    \bs M = \bs I - \bs P = \bs I - \bs X(\bs X^\intercal \bs X )^{-1}\bs X^\intercal
\end{equation}
such that $\bs M$ applied to an outcome vector $\bs y$ yields
\begin{equation}
    \bs M \bs y = \bs y -  \bs X(\bs X^\intercal \bs X )^{-1}\bs X^\intercal \bs y = \bs y - \bs P \bs y = \bs y - \bs X \bs{ \widehat{\beta}} = \bs{\hat{\epsilon}}.
\end{equation}

\noindent \textbf{Theorem:} \\ The FWL theorem states that equivalent to estimating 
\begin{equation}
    \bs y = \bs X_1 \bs{\widehat{\beta}_1} + \bs X_2 \bs{\widehat{\beta}_2} + \bs{\widehat{\epsilon}}
\end{equation}
we would obtain mathematically identical regression coefficients $\bs{\widehat{\beta}_1}$ and $\bs{\widehat{\beta}_2}$ if we separately estimated 
\begin{align}
    \bs M_2 \bs y & = \bs M_2 \bs X_1 \bs{\widehat{\beta}_1} + \bs{\hat{\epsilon}}  \\
    \text{and} & \nonumber \\ 
    \bs M_1 \bs y & = \bs M_1 \bs X_2 \bs{\widehat{\beta}_2} + \bs{\hat{\epsilon}}
\end{align}
where $\bs M_1$ and $\bs M_2$ correspond to the data partitions $\bs X_1$ and $\bs X_2$. 
\\~\\
\noindent \textbf{Proof of Theorem:} \\
This proof is based on the original papers \cite{frisch1933partial,lovell1963seasonal}.
Given 
\begin{equation}
    \bs y = \bs X_1 \bs{\widehat{\beta}_1} + \bs X_2 \bs{\widehat{\beta}_2} + \bs{\widehat{\epsilon}}
\end{equation}
left-multiply by $\bs M_2$, so we obtain
\begin{equation}
    \bs M_2 \bs y = \bs M_2 \bs X_1 \bs{\widehat{\beta}_1} + \bs M_2 \bs X_2 \bs{\widehat{\beta}_2} + \bs M_2 \bs{\widehat{\epsilon}}.
\end{equation}
We obtain from equation \eqref{eq_residual_maker} that
\begin{equation}
    \bs M_2 \bs X_2 \bs{\widehat{\beta}_2} = (\bs I - \bs X_2(\bs X_2^\intercal \bs X_2)^{-1}\bs X_2^\intercal)\bs X_2 \bs{\widehat{\beta}_2} = \bs X_2 \bs{\widehat{\beta}_2} - \bs X_2 \bs{\widehat{\beta}_2} = \bs 0.
\end{equation}
Finally, $ \bs M_2 \bs{\hat{\epsilon}} = \bs{\hat{\epsilon}}$. $\bs X_2 $ is orthogonal to $\bs \epsilon$ by construction of the OLS regression. Therefore, the residualized residuals are the residuals themselves.
Which leaves us with
\begin{equation}
    \bs M_2 \bs y = \bs M_2 \bs X_1 \bs{\widehat{\beta}_2} + \bs{\hat{\epsilon}} \qed. 
\end{equation}
The same goes through for $\bs M_1$ by analogy.

\subsection{$\mathbb{E}[\bs y | \bs t, \bs C]$, where $\bs t \perp \bs C$, no $\bs Z$}
In the simplest case assume there was no confounding text. Our observational data only consist of outcome $\bs y$, our treatment variable $\bs t$ and other potential confounding variables $\bs C$.
The conditional expectation function is $\mathbb{E}[\bs y | \bs t, \bs C]$. We can estimate it via one joint regression as
\begin{equation}\label{eq_no_t_perp_c}
    \bs y = \bs t \omega_t + \bs C \omega_C + \epsilon_0.
\end{equation}
Now, assuming that the linearity assumption is correct, the fact that $\bs t \perp \bs C$ implies that $\bs C $ is not actually a confounder in this setup. We would obtain the exact same regression coefficient estimates for $\omega_t$ and $\bs \omega_C$ if we followed a two-step process, in which we first regress $\bs y$ on $\bs t$
\begin{equation}\label{eq_no_t_perp_c_1}
    \bs y = \bs t \omega_t +  \epsilon_1.
\end{equation}
\begin{equation}\label{eq_no_t_perp_c_1}
    \bs y = \bs C \omega_C +  \epsilon_1.
\end{equation}
This is holds only true, if and only if $\bs t \perp \bs C$. Because in this case, $\bs t$ and $\bs C$ are already orthorgonal to each other. They already fulfill the requirements of the FWL and therefore such two-step process would yield mathematically equivalent regression coefficients $\bs \omega$ to the joint estimation in equation \eqref{eq_no_t_perp_c}. Put in terms of the conditional expectations, given linearity, $\mathbb{E[\bs y|\bs t, \bs C}] = \mathbb{E[\bs y|\bs t}] + \mathbb{E[\bs y|\bs C}] $, since $\bs t$ and $\bs C$ are uncorrelated and therefore $\bs C$ is not an actual confounder under the linear CEF setup.

\subsection{$\mathbb{E}[\bs y | \bs t, \bs C]$, where $\bs t \not\perp \bs C$, no $\bs Z$}
In this case, $\bs t \not\perp \bs C$. We now have $\mathbb{E[\bs y|\bs t, \bs C}]\neq \mathbb{E[\bs y|\bs t}]$ in the linear CEF setup, since $\bs C$ is a confounder. However, according to the FWL, we can still conduct separate stage regressions and obtain mathematically equivalent regression coefficients $\bs \omega$ if we residualize outcomes and regressors on all regressors that are not part of the partial regression.
We can estimate
\begin{equation}
    \bs M_C \bs y = \bs M_C \bs t \bs{\widehat{\omega}_t} + \bs{\hat{\epsilon}_1}
\end{equation}
and
\begin{equation}
    \bs M_t \bs y = \bs M_t \bs C \bs{\widehat{\omega}_C} + \bs{\hat{\epsilon}_2}
\end{equation}
and the obtained estimates $\widehat{\omega_t}$ and  $\bs{\widehat{\omega}_C}$ will be equivalent to those obtained from the joint estimation.

\subsection{$\mathbb{E}[\bs y | \bs t, \bs C, \bs Z ]$, where $\bs t \not\perp \bs C$,$\bs Z$}
We now consider the case where part (or all) of our confounders are text or where text is a proxy for otherwise unobserved confounders. The joint estimation would be
\begin{equation}
    \bs y = \bs t \bs{\widehat{\omega}_t} +
    \bs C \bs{\widehat{\omega}_C} + 
    \bs Z \bs{\widehat{\omega}_Z} +
    \bs{\hat{\epsilon}}
\end{equation}
where $\bs Z$ itself is obtain through supervised learning via text representation function
$$
\bs Z = g(\bs W, \bs y, \bs t, \bs C; \Theta).
$$
We therefore cannot decompose this joint estimation into separate parts. As long as the text features $\bs Z$ are correlated with the outcome and the other covariates, we would need to apply the orthogonalization via the respective $\bs M$ matrices for each partical regression. Since $\bs Z$ needs to be estimated itself (it is `estimated data'), we cannot residualize on $\bs Z$ though. Nor can $\bs Z$ be residualized on the other covariates. A separate-stage approach will therefore lead to biased estimates of $\bs \omega$.

\section{Regression Model}\label{section:app_reg_model}

Due to the conjugacy of the Normal-Inverse-Gamma prior, the posterior distribution of the regression parameters conditional on $\bs{A}$ has a known Normal-Inverse-Gamma distribution:
\begin{align}\label{app:omega_posterior}
p(\bs{\omega},\sigma^{2} | & \bs{y}, \bs{A})  \propto p(\bs{\omega} | \sigma^{2},\bs{y} ,\bs{A} )p (\sigma^{2}\mid \bs{y}, \bs{A}) = \mathcal{N}\left(\bs{\omega} | \bs{m}_{n},\sigma ^{2}\bs{S}_{n}^{-1}\right) {\mathcal{IG}}\left(\sigma^2 | a_{n},b_{n}\right)
\end{align}

\noindent where $\bs{m}_{n}$, $\bs{S}_{n}$, $a_{n}$ and $b_{n}$ follow standard updating equations for a Bayesian Linear Regression (Bishop 2006)
%\small
\begin{align}
& \bs{m}_{n}=(\bs{A}^\intercal \bs{A} +\bs{S}_0)^{-1}(\bs{S}_0 \bs{m}_0+\bs{A} ^{\intercal}\bs{y}) \\ 
& \bs{S}_{n}=(\bs{A} ^{\intercal}\bs{A} + \bs{S}_0) \\
& a_{n}=a_{0}+ N/2 \\
& b_{n}=b_{0}+ \left( \bs{y}^{\intercal}\bs{y} + \bs{m}_{0}^{\intercal} \bs{S}_{0}\bs{m}_{0} - \bs{m}_{n}^{\intercal}\bs{S}_{n}\bs{m}_{n}\right)/2.
\end{align}
\normalsize

\section{Topic Model} \label{section:app_topic_model}

\subsection{Gibbs-EM algorithm}
\subsubsection{Sampling distribution for $z$}

The probability of a given word $w_{d,n}$ being assigned to a given topic $k$ (such that $z_{d,n}=k$), conditional on the assignments of all other words (as well as the model's other latent variables and the data) is 
\begin{align}
p(z_{d,n} = k | \bs{Z}_{-(d,n)}, \bs{W},\bs{X},\bs{y}, \bs{\omega}, \sigma^2),
\end{align}
where $\bs{Z}_{-(d,n)}$ are the topic assignments for all words apart from $w_{d,n}$. By the conditional independence properties implied by the graphical model, we can split this joint posterior into
\begin{align}
\label{sLDA_cov_posterior_form}
p(\bs{Z}|\bs{W},\bs{X},\bs{y}, \bs{\omega}, \sigma^2) \propto p(\bs{Z}|\bs{W}) p(\bs{y}|\bs{Z}, \bs{X}, \bs{\omega}, \sigma^2).
\end{align}
As topic assignments within one document are independent from topic assignments in all other documents, the sampling equation for the $n$th word in document $d$ should only depend it's own response variable, $y_d$, such that
%\small
\begin{align}
\label{sLDA_cov_sampling_form}
p(z_{d,n} = k | \bs{Z}_{-(d,n)}, \bs{W}, \bs{X}, \bs{y}, \bs{\omega}, \sigma^2)  \propto
p(z_{d,n} = k | \bs{Z}_{-(d,n)}, \bs{W}) p(y_d|z_{d,n}=k, \bs{Z}_{-(d,n)}, \bs{x}_d, \bs{\omega}, \sigma^2).
\end{align}
\normalsize
The first part of the RHS expression is just the sampling distribution of a standard LDA model, so it can be expressed in terms of the count variables $\bs{s}$ (the topic assignments across a document) and $\bs{m}$ (the assignments of unique words across topics over all documents). $s_{d,k}$ measures the total number of words in document $d$ assigned to topic $k$ and $s_{d,k,-n}$ the number of words in document $d$ assigned to topic $k$, except for word $n$. Analogously, $m_{k,v}$ measures the total number of times term $v$ is assigned to topic $k$ across all documents and $m_{k,v,-(d,n)}$ measures the same, but excludes word $n$ in document $d$.
%\small
\begin{align}
p(z_{d,n} = k | \bs{Z}_{-(d,n)}, \bs{W}) \propto
(s_{d,k, -n } + \alpha) \frac{m_{k,v, -(d,n)} +\eta}{\sum_v m_{k,v, -(d,n)} +V \eta}.
\end{align}
\normalsize

\subsubsection{Regression}
Given that the residuals are Gaussian, the probability of the response variable for a given document $d$ is
%\small
\begin{align}
p(y_d | \bs{z}_d, \bs{x}_d, \bs{\omega}, \sigma^2) = \frac{1}{\sqrt{2\pi \sigma^2}} \exp\left\{-\frac{(y_d-\bs{\omega}^\intercal \bs{a}_d)^2}{2\sigma^2}\right\}. 
\end{align}
\normalsize
We can write this in a convenient form that preserves proportionality with respect to $z_{d,n}$ such that it depends only on the data and count variables used in the other two terms.
First, we split the $\bs{x}_d$ features into those that are interacted, $\bs{x}_{1,d}$, and those that are not, $\bs{x}_{2,d}$. The generative model for $y_d$ is then
%\small
\begin{align}
y_d \sim \mathcal{N}(\bs{\omega}_z^\intercal \bs{\bar{z}}_{d} + \bs{\omega}_{zx}^\intercal (\bs{x}_{1,d} \otimes \bs{\bar{z}}_d) + \bs{\omega}_x^\intercal \bs{x}_{2,d}, \sigma^2).
\end{align}
\normalsize
where $\otimes$ is the Kronecker product. Noting that $\bs{X}$ is observed, so we can think of this as a linear model with document-specific regression parameters. Define $\bs{\tilde{\omega}}_{z,d}$ as a length $K$ vector such that
%\small
\begin{align}
\tilde{\omega}_{z,d,k} = \omega_{z,k} + \bs{\omega}_{zx,k}^\intercal \bs{x}_{1,d}.
\end{align}
\normalsize
Noting that $\bs{\tilde{\omega}}_{z,d}^\intercal \bs{\bar{z}}_d = \frac{\bs{\tilde{\omega}_{z,d}}^\intercal}{N_d}(\bs{s}_{d,-n} + \bs{s}_{d,n})$, the probability density of $y$ conditional on $z_{d,n} = k$ is therefore proportional to
\begin{align}
& p(y_d | z_{d,n} = k, \bs{z}_{-(d,n)}, \bs{x}_d, \bs{\omega}, \sigma^2) \propto \nonumber \\ 
& \exp\left\{\frac{1}{2 \sigma^2} \left( \frac{2 \tilde{\omega}_{z,d,k}}{N_d} \left(y_d - \bs{\omega}_x^\intercal \bs{x}_d - \frac{\bs{\tilde{\omega}}_{z,d}^\intercal}{N_d}\bs{s}_{d,-n}\right) - \left(\frac{\tilde{\omega}_{z,d,k}}{N_d} \right) ^2 \right) \right\}.
\end{align}
This gives us the sampling distribution for $z_{d,n}$ stated in equation \eqref{sLDA_cov_sampling_form}: a multinomial distribution parameterised by 
\begin{align} 
& p(z_{d,n} = k | \bs{Z}_{-(d,n)}, \bs{W},\bs{X},\bs{y}, \alpha, \eta, \bs{\omega}, \sigma^2) \propto \nonumber \\ 
& (s_{d,k, -n } + \alpha) \frac{m_{k,v, -(d,n)} +\eta}{\sum_v m_{k,v, -(d,n)} +V \eta} \nonumber \\ 
& \exp\left\{\frac{1}{2 \sigma^2} \left( \frac{2 \tilde{\omega}_{z,d,k}}{N_d} \left(y_d - \bs{\omega}_x^\intercal \bs{x}_{2,d} - \frac{\bs{\tilde{\omega}}_{z,d}^\intercal}{N_d}\bs{s}_{d,-n}\right) - \left(\frac{\tilde{\omega}_{z,d,k}}{N_d} \right) ^2 \right) \right\}.
\end{align}
\normalsize
This defines for each $k \in \{1,...,K\}$ the probability that $z_{d,n}$ is assigned to that topic. These $K$ probabilities define the multinomial distribution from which $z_{d,n}$ is drawn.

\subsubsection{$\theta$ and $\beta$}

Given topic assignments $z$, we can recover the latent variables $\theta$ and $\beta$ from their predictive distributions via
\begin{align}
\hat{\theta}_{d,k} = \frac{s_{d,k}+\alpha}{\sum_k(s_{d,k}+\alpha)}    
\end{align}
and
\begin{align}
\hat{\beta}_{k,v} = \frac{m_{k,v} + \eta}{\sum_v(m_{k,v} + \eta)}.  
\end{align}

\subsubsection{Observations without documents} \label{section_obs_wo_docs}

A straightforward extension allows for some observations to be associated with an $\bs{X}$ and $\bs{y}$, but no document. This is often the case in a social science context, for example time-series may be associated with documents at irregular intervals. If an observation is not associated with any documents, the priors on the document topic distributions suggest that the topic assignment for topic $K$ is set to $\alpha_k/\sum_k\alpha_k$. These observations may still be very useful in estimating the relationship between $\bs{X}$ and $\bs{y}$ so they are worth including in the estimation.

\subsubsection{Multiple paragraphs}
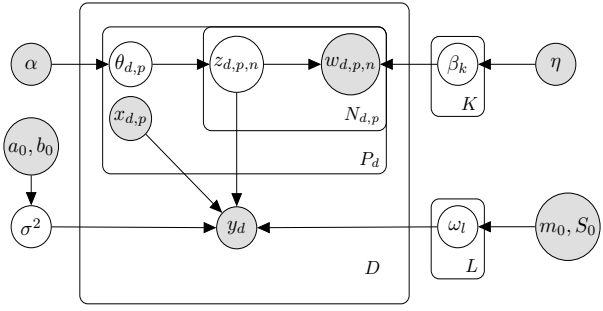
\begin{figure}
	\centering
	\resizebox{8cm}{4cm}{
	\begin{tikzpicture}
	% nodes
	\node[obs] (alpha) {$\alpha$}; %
	\node[latent,right=of alpha ] (theta) {$\theta_{d,p}$}; %
	\node[latent,right=of theta ] (z) {$z_{d,p,n}$}; %
	\node[obs, right=of z] (w) {$w_{d,p,n}$};%
	\node[latent,right=of w,yshift= 0cm,fill] (beta) {$\beta_k$}; %
	\node[obs,right=of beta,xshift=0cm,fill] (eta) {$\eta$}; %
	\node[obs, below=of z, yshift = -0.9cm, ] (y) {$y_d$};%
	\node[obs, below=of theta, yshift = 0.85cm, ] (xp) {$x_{d,p}$};%
	\node[latent,below=of beta,yshift=-1cm] (omega) {$\omega_l$}; %
	\node[obs,right=of omega,xshift=0cm] (m) {$m_0, S_0$}; %
	\node[latent,below=of alpha,yshift=-1.0cm] (sigma) {$\sigma^2$}; %
	\node[obs,above=of sigma,yshift=-0.5cm] (sigma_priors) {$a_0, b_0$}; %
	%\draw (0,1.69) circle(.38cm);
	% plate
	\plate [inner sep=.5cm,yshift=.0cm] {plateD} {(theta)(z)(w)(y)(xp)} {$D$}; %
	\plate [inner sep=.1cm,yshift=0cm] {plateP} {(theta)(z)(w)(xp)} {$P_d$}; %
	\plate [inner sep=.1cm,yshift=.0cm] {plateN} {(z)(w)} {$N_{d,p}$}; %	
	\plate [inner sep=.1cm,yshift=.0cm] {plateK} {(beta)} {$K$}; %
	\plate [inner sep=.1cm,yshift=.0cm] {plateM} {(omega)} {$L$}; %				
	% edges
	\edge {alpha} {theta} 
	\edge {theta} {z} 
	\edge {z} {w}  
	\edge {beta} {w}  
	\edge {eta} {beta}  
	\edge {xp} {y}
	\edge {z} {y}
	\edge {omega} {y}
	\edge{m} {omega}
	\edge{sigma} {y}
	\edge{sigma_priors} {sigma}
	\end{tikzpicture}
	}
	\caption{Graphical model for BTR with multiple documents per observation}
	\label{fig:btr_para_new}
\end{figure}

If, as is often the case in the context of social science applications, we have relatively few observations but the documents associated with those observations are relatively long, we can exploit the structure of the documents by estimating the model at a paragraph level. Splitting up longer documents into paragraphs brings one of the key advantages of topic modelling to the fore: that the same word can have different meanings in different contexts. For example, the word ``increase" might have quite a different meaning if it is in a paragraph with the word ``risk" than if it is alongside ``productivity". Treating the entire document as a single bag of words makes it hard for the model to make this distinction.
\\~\\
If there are observations with multiple documents, we can treat these as $P_d$ separate paragraphs of a combined document, indexed by $p$, each with an independent $\bs{\theta}_p$ distribution over topics. These paragraphs may also have different associated $\bs{x}_{d,p}$ that interact with the topics, for example we may wish to interact topics with a paragraph specific sentiment score, but the response variable $y_d$ is common to all paragraphs in the same document and the M-step estimated at the document level. Figure \ref{fig:btr_para_new} shows the extended graphical model.

If $\bs{x_{d,p}}$ only enters linearly into the regression then some document-level average will have to be used and this transformation can be performed prior to estimation, converting it into an $\bs{x}_{1,d}$, and so the algorithm will remain unchanged. However, if any of the $\bs{x}_{d,p}$ variables are interacted with $\bs{\bar{z}}_{d,p}$ then we may wish for this interaction to be at the paragraph level. For example, if we think that a topic might have a different effect depending on the sentiment of the surrounding paragraph. In this case, we still need to aggregate the interaction to the document level, but aggregate after interacting rather than interacting after aggregating. We therefore define
\begin{align}
\overline{\bs{x}_{d,p}\otimes \bs{z}_{d,p}} = \frac{1}{N_d}\sum_{p\in[P_d]}\sum_{n\in[N_{d,p}]}\left[\bs{x}_{d,p} \otimes  \bs{s}_{d,p,n} \right]
\end{align}
where $[N]$ denotes the set of integers $\{1,...,N\}$ and $\otimes$ represents the Kronecker product.
The design matrix $\bs{A}$ is then 
\begin{align}
\bs{A} = \left[\begin{matrix}
\bs{\bar{z_{1}}}& \overline{ \bs{x}_{1,1,p} \otimes \bs{z}_{1,p}} & \bs{x}_{2,1} \\
\vdots & \vdots & \vdots \\
\bs{\bar{z}}_{1}& \overline{\bs{x}_{1,d,p} \otimes \bs{z}_{d,p} } & \bs{x}_{2,d} \\
\vdots & \vdots & \vdots \\
\bs{\bar{z}}_{1}& \overline{\bs{x}_{1,D,p} \otimes \bs{z}_{D,p}} & \bs{x}_{2,D} \\
\end{matrix}\right]
\end{align}
and the predictive model for $y_d$ will be
\begin{align}
\bs{y} \sim \mathcal{N}(\bs{A}\bs{\omega}, \sigma^2) \ \  \text{where} \ \ \bs{\omega} = (\bs{\omega}_z, \bs{\omega}_{zx}, \bs{\omega}_{x}).
\end{align}
The simplest way to aggregate from paragraphs to documents is simply to give each word in the document equal weight as above. This will mean that longer paragraphs have greater weight than shorter ones.
\\~\\
As before, we can collapse out the latent variables $\bs{\theta}$ and $\bs{\beta}$ so that we only need to sample for the topic assignments $\bs{z}$ in an E-step and then for $\bs{\omega}$ and $\sigma^2$ in an M-step.
\\~\\
In the E-step, we need to sample from the conditional posterior for the topic assignment of each word
\begin{align}
\Pr[z_{d,p,n} = k | \bs{Z}_{d,-(p,n)}, \bs{W}, \alpha, \eta, \bs{y}, \bs{X}, \bs{\omega}, \sigma^2].
\end{align}
By the conditional independence properties of the graphical model, we can split this into $p(\bs{Z}|\bs{W},\alpha,\eta)$ and $p(\bs{y}|\bs{Z},\bs{X},\bs{\omega},\sigma^2)$. The sampling equation for the $n$th token in the $p$th paragraph of the $d$th document $d$ will have the form
\begin{equation}
\begin{split}
\Pr[z_{d,p,n} = k | \bs{Z}_{d,-(p,n)}, \bs{W}, \alpha, \eta, \bs{y}, \bs{X}, \bs{\omega}, \sigma^2] \propto & \\ \Pr[z_{d,p,n} = k | \bs{Z}_{d,p,-(n)}, \bs{W}, \alpha, \eta] & \ \ \times \ \ \Pr[y_d|z_{d,p,n}=k, \bs{Z}_{d,-(p,n)}, \bs{x}_d, \omega, \sigma^2].
\end{split}
\end{equation}
The topic assignment each document is independent, but there are dependencies across paragraphs. Crucially, these paragraphs have are independent with respect to $\bs{\theta}$, so $p(\bs{Z}|\bs{W},\alpha,\eta)$ is paragraph specific.
\begin{equation}
\Pr[z_{d,p,n} = k | \bs{Z}_{d,p-(n)}, \bs{W}, \alpha, \eta] \propto (s_{d,p,k, -n } + \alpha) \frac{m_{k,v, -(d,p,n)} +\eta}{\sum_v m_{k,v, -(d,p,n)} +V \eta}.
\end{equation}
However, the regression part is at the document level to $p(\bs{y}|\bs{Z},\bs{X},\bs{\omega},\sigma^2)$ will condition on all the paragraphs in a given document. Given that the residuals are Gaussian, the probability of the outcome variable for a given document $d$ is 
\begin{align}
p(\bs{y}_d | \bs{z}_d, \bs{x}_d, \bs{\omega}, \sigma^2) = \frac{1}{\sqrt{2\pi \sigma^2}} \exp\left[-\frac{(y_d-\bs{\omega}_z^\intercal\bs{\bar{z}}_d - \bs{\omega}_{zx}^\intercal(\overline{\bs{x}_{1,d,p} \otimes \bs{z}_{d,p}}) - \bs{\omega}_x^\intercal \bs{x}_{2,d})^2}{2\sigma^2}\right].
\end{align}
We can write this in a convenient form that preserves proportionality with respect to $z_{d,p,n}$ such that it depends only on the data and count variables used in the other two terms and the document-wide counts. First we can break the prediction for $y_d$ into the section that depends on paragraph $p$ and the section that depends on other paragraphs and document wide $\bs{x}_{1,d}$.
\begin{equation}%\small
\begin{split}
y_d -  \bs{\omega}_z^\intercal\bs{\bar{z}}_d -  \bs{\omega}_{zx}^\intercal(\overline{\bs{x}_{1,d,p} \otimes \bs{z}_{d,p}})  = &  \left(y_d - \bs{\omega}_x^\intercal \bs{x}_{2,d} -  
\frac{\bs{\omega}_{z}^\intercal}{N_d}\bs{s}_{d,-p} - \frac{\bs{\omega}_{zx}^\intercal}{N_d} \sum_{q\in\{[P_d]\backslash p\}}\left[\bs{x}_{1,d,q}\otimes \bs{s}_{d,q}\right]\right) \\ 
& - \left(\frac{\bs{\omega}_{z}^\intercal}{N_d}(\bs{s}_{d,p,-n} + \bs{s}_{d,p,n}) - 
\frac{\bs{\omega}_{zx}^\intercal}{N_d} \bs{x}_{1,d,p}\otimes (\bs{s}_{d,p,-n} + \bs{s}_{d,p,n}) \right)
\end{split}
\end{equation} 
where $N_d$ is the total number of words in the \textit{document}. 
\\~\\
Define $\hat{y}_{d,-p}$ as the predicted $y_d$ without paragraph $p$,
\begin{align}
\hat{y}_{d,-p} = \bs{\omega}_x^\intercal \bs{x}_{2,d} +
\frac{\bs{\omega}_{z}^\intercal}{N_d}\bs{s}_{d,-p} + \frac{\bs{\omega}_{zx}^\intercal}{N_d} \sum_{q\in\{[P_d]\backslash p\}}\left[\bs{x}_{1,d,q}\otimes \bs{s}_{d,q}\right].
\end{align}
We then have a predictive distribution that depends only on paragraph $p$.
\begin{align}
y_d \sim \mathcal{N}\left(\hat{y}_{d,-p}  - 
\frac{\bs{\omega}_{z}^\intercal}{N_d}(\bs{s}_{d,p,-n} + \bs{s}_{d,p,n}) - 
\frac{\bs{\omega}_{zx}'}{N_d} \bs{x}_{1,d,p}\otimes (\bs{s}_{d,p,-n} + \bs{s}_{d,p,n}), \sigma^2\right).
\end{align}
We can then follow the same steps as for the single paragraph document case to derive the third term in the sampling distribution, defining $\bs{\tilde{\omega}}_{z,d,p,k} = \bs{\omega}_{z,k} + \bs{\omega}_{zx,k}'\bs{x}_{1,d,p}$ analogously to $\bs{\tilde{\omega}}$ defined for the single paragraph case.
\\~\\
This gives us the sampling distribution for $z$, which is a Multinomial parameterised by 
\begin{equation}
\small
\begin{split}
\Pr[z_{d,n} = k | \bs{Z}_{-(d,n)}, \bs{W}, \bs{y}, \alpha, \eta, \bs{\omega}, \sigma^2] \propto & 
(s_{d,p,k, -n } + \alpha) \frac{m_{k,v, -(d,p,n)} +\eta}{\sum_v m_{k,v, -(d,p,n)} +V \eta} \\ &
\exp\left[\frac{1}{2 \sigma^2} \left( \frac{2 \tilde{\omega}_{z,d,p,k}}{N_d} \left(y_d - \hat{y}_{d,-p}- \frac{\bs{\tilde{\omega}}_{z,d,p}'}{N_d}\bs{s}_{d,-n}\right) - \left(\frac{\tilde{\omega}_{z,d,p,k}}{N_d} \right)\right) ^2 \right].
\end{split}
\end{equation}
~\\
In the M-step we can then still use the average $\bar{z}_{d,p}$ estimated in the E-step, but we need to weight each paragraph by the number of words in that paragraph to be consistent with the E-step,
\begin{align}
\bar{z}_{d} = \frac{1}{N_d}\sum_{p\in[P_d]}\left[ N_{d,p} \bar{z}_{d,p}\right]
\end{align}
\begin{align}
(\overline{x_{1,d,p} \otimes z_{d,p}}) = \frac{1}{N_d}\sum_{p\in[P_d]}\left[ N_{d,p} x_{1,d,p} \otimes z_{d,p}\right].
\end{align}

%-------------------------------
%-------------------------------
%-------------------------------

\section{Synthetic Data Experiments}

\begin{figure}
    \centering
    \includegraphics[width = 0.525 \linewidth]{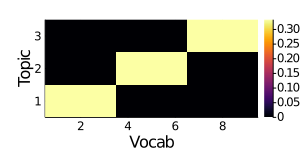}
    \caption{Ground truth topic distribution for synthetic documents.}
    \label{fig:true_topics}
\end{figure}

Figure \ref{fig:true_topics} shows the topic-vocabulary distribution from which the synthetic documents are generated. 

Table \ref{app:synth_hyper_table} shows the hyperparameter settings used in the synthetic data section. We observed that the settings of the prior did hardly effect results, given the strong signal in the synthetic dataset.
\begin{table}[H]
  \centering
  \caption{Synthetic example hyperparameters}
        \resizebox{0.6\linewidth}{!}{\begin{tabular}{lccccccccc}
          & K & $\alpha$ & $\eta$ & $\mu_\text{ntm}$ & $\sigma_\text{ntm}$ & $a_0$ & $b_0$ & $m_0$ & $S_0$ \\
    \midrule
    LDA   & 3 & 1.0 & 1.0 & -     & -     & -     & -     & -     & - \\
    sLDA  & 3 & 1.0 & 1.0 & -     & -     & -     & -     & -     & - \\
    BPsLDA & 3 & 1.0 & 1.0 & -     & -     & -     & -     & -     & - \\
    BTR  & 3 & 1.0 & 1.0 & -     & -     & 0.2 & 4 & 0     & 2 \\
    \bottomrule
    \end{tabular}}
  \label{app:synth_hyper_table}%
\end{table}%

\section{Semi-Synthetic Data Experiments}
\label{section:app_semisynth}
\begin{figure}[H]
    \centering
    \includegraphics[width = 0.6 \linewidth]{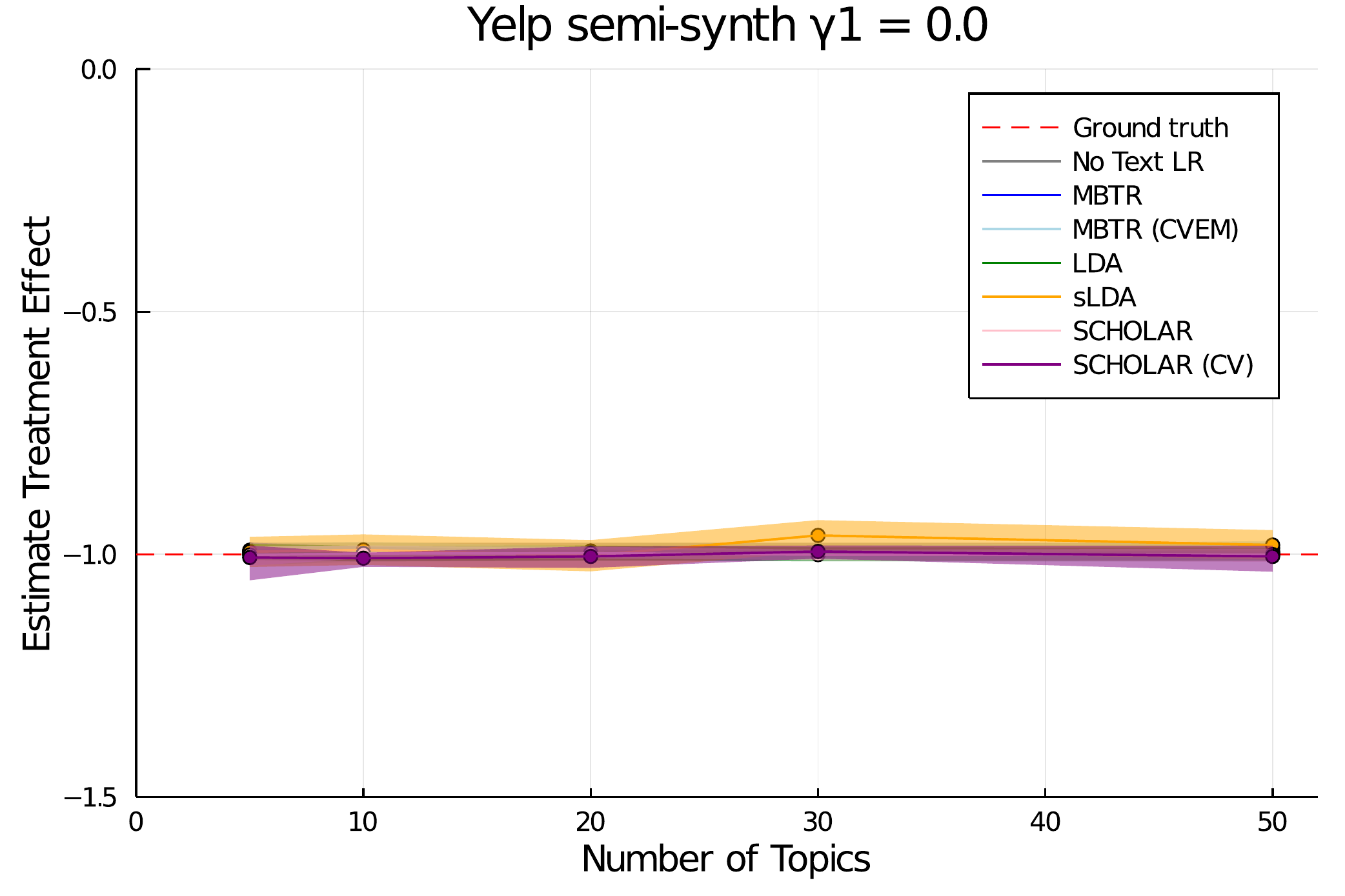}
    \caption{Without correlation between confounders and treatments, the regression can be dissected into two separate parts (supervised topic estimation and regression weight estimation of the non-text features) without inducing bias in the estimators,  as described in the section on the Frisch-Waugh-Lovell theorem. In such a case, all models manage to recover the ground truth.}
    \label{fig:app_true_topics_semi_synth_yelp}
\end{figure}

\section{Real-World Datasets and Data Pre-Processing}\label{section_app_real_world_data}
The\textbf{ Yelp dataset} contains over $8$ million customer reviews of businesses, which we restrict to reviews for businesses in Toronto. The \textbf{Booking dataset} contains around $500,000$ hotel reviews.  For both datasets, we randomly sample $50,000$ observations and randomly select $75\%$ in Yelp, $80\%$ in Booking of our sample for training, holding out the remainder for testing. We then further split the training set equally for training in the E-step and validation in the M-step. The features are normalized on the training data statistics and the response variable is de-meaned. We do this because the $K$ topic features sum to one and therefore implicitly already add a constant to the regression \citep{blei_mcauliffe2008supervised}.
We preprocess the text corpora by removing stopwords and then tokenizing and stemming the data.

\begin{table}[H]
    \centering
    \caption{Summary statistics of the review datasets}
    \label{tab:app_corpus_stats}
    \begin{tabular}{@{\extracolsep{0pt}} lcccccc}
    \toprule
    \textbf{Statistics}  & \#train & \#val & \#test  &  \#vocab &  \#max words & \#avg words \\
    \midrule
    \textbf{Yelp} & 18,750  &  18,750  &  12,500 & 24,680 & 572 & 61.2 \\
    \textbf{Booking} & 20,000  &  20,000  &  10,000 & 6,968 & 305 & 18.7 \\
    \bottomrule
    \end{tabular}
\end{table}

The Booking.com dataset allows consumers to enter the positive and negative parts of their reviews in separate boxes. We combine these two reviews for all our exercises, but we do use information on the word count in each of these sections (see below). 
\\~\\
For the prediction exercises in Section \ref{section:emp_experiments}, we use the number of stars associated with each review as the target variable. We also use the numerical metadata described in Table \ref{tab:num_covs}as covariates.

\begin{table}[H]
    \centering
    \caption{Numerical covariates for prediction experiments}
    \label{tab:num_covs}
    \resizebox{\linewidth}{!}{
    \begin{tabular}{@{\extracolsep{0pt}} lcc}
    \toprule
    \textbf{Dataset}  & Variable & Description \\
    \midrule
    \multirow{3}{*}{Yelp} & stars\_av\_u & historic avg. rating by user \\
    & stars\_av\_b & historic avg. rating of business \\
    & sentiment & \textit{Harvard Inquirer} sentiment score \\
    \midrule
    \multirow{3}{*}{Booking} & Average\_Score & historical avg. hotel score \\
    & Review\_Total\_Negative\_Word\_Counts & total number of words in the negative part of review \\
    & Review\_Total\_Positive\_Word\_Counts & total number of words in the positive part of review \\
    & Total\_Number\_of\_Reviews\_Reviewer\_Has\_Given & total num of reviews by customer \\
    & Total\_Number\_of\_Reviews & total num of reviews of hotel \\
    \bottomrule 
    \end{tabular}}
\end{table}

For the semi-synthetic exercise on the Booking data, we construct 
$$
pos\_prop_i = \frac{Review\_Total\_Positive\_Word\_Counts_i}{Review\_Total\_Positive\_Word\_Counts_i + Review\_Total\_Negative\_Word\_Counts_i}
$$
This variable is correlated with the treatment ($Average\_Score_i$) and with the outcome, and so the text can act as a confounder. 

\section{Real-World Data Experiments}
\label{section:app_realworld_experiments}
\subsection{Empirical data evaluation across different K}\label{app_empirical_data_results}
\begin{table}[H]
\centering
\caption{Mean $pR^2$ and perplexity over 20 runs per model, standard deviation in brackets}
    \label{tab:pR_perplexity}
\resizebox{\linewidth}{!}{
    \begin{tabular}{@{\extracolsep{0pt}} lcccccccc}
    \toprule
    \textit{Dataset} & \multicolumn{4}{c}{Booking} & \multicolumn{4}{c}{Yelp} \\
    \midrule
    \textit{K}  & 10 & 20 & 30  &  50 &  10 & 20 & 30 & 50 \\
    \midrule
    \multicolumn{9}{c}{$pR^2$ (higher is better)} \\
    \midrule
    LDA+LR & 0.400 (0.003) & 0.410 (0.004) & 0.417 (0.005) & 0.426 (0.003) & 0.498 (0.005) & 0.530 (0.009) & 0.561 (0.010) & 0.586 (0.006) \\
    GSM+LR & 0.387 (0.003) & 0.390 (0.004) & 0.389	(0.006) & 0.386	(0.004) & 0.502	(0.013) & 0.505 (0.011) & 0.503 (0.008) & 0.495 (0.004) \\
    LR+sLDA & 0.416 (0.007) & 0.426 (0.003) & 0.430 (0.004) & 0.432 (0.002) & 0.533 (0.007) & 0.564 (0.003) & 0.567 (0.006) & 0.571 (0.002) \\
    LR+BPsLDA & 0.394 (0.004) & 0.396 (0.005) & 0.400 (0.005) & 0.419 (0.009) & \textbf{0.593 (0.003)} & \textit{0.597} (0.002) & \textit{0.597} (0.002) & \textit{0.603} (0.002) \\
    rSCHOLAR & \textbf{0.494} (0.005) & \textbf{0.495} (0.003) & \textbf{0.495} (0.003) & \textbf{0.494} (0.004) & 0.520 (0.02) & 0.548 (0.02) & 0.563 (0.01) & 0.571 (0.01) \\
    \textbf{BTR} & \textit{0.439} (0.008) & \textit{0.447} (0.005) & \textit{0.453} (0.003) & \textit{0.454} (0.003) & \textit{0.586} (0.007) & \textbf{0.615} (0.006) & \textbf{0.627} (0.004) & \textbf{0.630} (0.001) \\
    \bottomrule
    \\[-1.8ex] 
    \multicolumn{9}{c}{Perplexity (lower is better)} \\
    \midrule
    LDA+LR & 538 (3) & 498 (2) & 476 (2) & 454 (1) & 1544 (5) & 1447 (4) & 1388 (4) & 1306 (4) \\
    GSM+LR & \textbf{371 (6)} & \textbf{359 (11)} & \textbf{356 (14)} & \textbf{369 (8)} & \textbf{1500 (52)} & \textit{1444} (29) & 1463 (21) & 1431 (34)  \\
    LR+sLDA & \textit{535} (2) & 491 (1) & \textit{463} (1) & \textit{436} (2) & 1544 (6) & \textit{1444} (6) & 1382 (5) & \textit{1294} (5) \\
    rSCHOLAR & 941 (134) & 1429 (163) & 2110 (396) & 5014 (1314) & 1744 (158) & 1918 (138) & 2216 (164) & 2814 (383) \\
    \textbf{BTR} & \textit{535} (2) & \textit{490} (1) & \textit{463} (2) & 437 (1) & \textit{1540} (5) & \textbf{1443 (4)} & \textbf{1379 (4)} & \textbf{1291 (5)}  \\
    \bottomrule
    \end{tabular}}
    \caption{Best model in \textbf{bold}. Second best model in \textit{italics}.}
\end{table}

We also tested sLDA+LR and a pure sLDA, which performed consistently worse so they are not included for the sake of brevity. For example, for $K=50$, sLDA+LR achieved $pR^2$ of $0.420$ and $0.564$ for Booking and Yelp respectively, compared to $0.432$ and $0.571$ for LR+sLDA. Standalone sLDA achieves $0.356$ and $0.526$ respectively.

\subsection{Model parametrisations} \label{app:model_params}

This section provides an overview over all used and tested hyperparameter settings across all models in our benchmark list. Table \ref{app:topic_hyper_paras} lists all hyperparameter settings pertaining to topic model components. Table \ref{tab:nn_hyper_paras} provides an overview over all used neural network hyperparameters. \ref{app:iter_hyperparas} summarises the iteration and stopping criteria for all models.

\begin{table}[H]
  \centering
  \caption{Topic model hyperparameters}
    \resizebox{\linewidth}{!}{\begin{tabular}{lccccccccc}
          & K & $\alpha$ & $\eta$ & $\mu_\text{ntm}$ & $\sigma_\text{ntm}$ & $a_0$ & $b_0$ & $m_0$ & $S_0$ \\
    \midrule
    LDA   & [10,20,30,50] & [0.1,0.5,1] & [0.001,0.01,0.1] & -     & -     & -     & -     & -     & - \\
    sLDA  & [10,20,30,50] & [0.1,0.5,1] & [0.001,0.01,0.1] & -     & -     & -     & -     & -     & - \\
    BPsLDA & [10,20,30,50] & [0.1,0.5,1] & [0.001,0.01,0.1] & -     & -     & -     & -     & -     & - \\
    BTR  & [10,20,30,50,100] & [0.1,\textbf{0.5},1] & [0.001,\textbf{0.01},0.1] & -     & -     & [0,1.5,\textbf{3},\textbf{4}] & [0,\textbf{2},4] & 0     & 2 \\
    GSM   & [10,20,30,50] & -     & -     & 0     & 1     & -     & -     & -     & - \\
    TAM   & 100   & -     & -     & 0     & 1     & -     & -     & -     & - \\
    \bottomrule
    \multicolumn{10}{c}{\textit{Bold parameter specifications were used for reported results in paper, unless stated otherwise. For Booking default $a_0=3$, for Yelp $a_0=4$.}} \\
    \end{tabular}}%
  \label{app:topic_hyper_paras}%
\end{table}%

\begin{table}[H]
  \centering
  \caption{Neural network hyperparameters}
    \resizebox{\linewidth}{!}{\begin{tabular}{ccccccccc}
          & HidLaySize &  &  & DropOut &  & HidLaySize & & nHidLayers \\
          & $\theta$ & BatchSize & LearnRate & KeepRate & EmbedSize &RNN & TAM-thresh & BPsLDA \\
    \midrule
    GSM   & 64    & 64    & 1.00E-03 & [0.5,0.8,1]* & -     & -     & -     & - \\
    TAM   & 64    & 64    & 1.00E-03 & 0.8   & 100   & 64    & 1/K   & - \\
    aRNN   & -     & 64    & 1.00E-03 & 0.8   & 100   & 64    & -     & - \\
    BPsLDA & -     & 1050  & 1.00E-02 & -     & -     & -     & -     & 10 \\
    \bottomrule
    \multicolumn{9}{c}{\textit{* best results (which occurred under no dropout) were reported in benchmarks}} \\
    \end{tabular}}%
  \label{tab:nn_hyper_paras}%
\end{table}%

\begin{table}[H]
  \centering
  \caption{Iteration parameters}
  \resizebox{\linewidth}{!}{\begin{tabular}{ccccccc}
          & E-step iters & M-step iters & max. EM-iters & burn-in & max. epochs & Gibbs iters (thinning) \\
    \midrule
    LDA   & - & -  & 50***  & 100    & - & 1000 (5) \\
    sLDA  & [100,250,500]** & 2500  & 50***  & 20    & - & -\\
    BTR & [100,250,500]** & 2500  & 50***  & 20    & - & - \\
    GSM   & -     & -     & -     & -     & 100*** & - \\
    TAM   & -     & -     & -     & -     & 100*** & -\\
    BPsLDA & -     & -     & -     & -     & 50*** & -\\
    \bottomrule
    \multicolumn{6}{c}{\textit{** no noticeable performance difference observed, therefore all results reported based on 100 E-step.}} \\
    \multicolumn{6}{c}{\textit{*** best model achieved substantially before max. iterations reached.}} \\
    \end{tabular}}%
  \label{app:iter_hyperparas}%
\end{table}%

\textbf{Further notes on benchmark model specifications:}\\~\\
{\textbf{For TAM and aRNN}, the sequence length in the RNN component (ie. the maximum number of words per document) is $305$ for Booking and $572$ for Yelp which corresponds to the longest review in each respective data set.  We therefore work with the full text of each review.
}

{\textbf{BPsLDA}} changes its behaviour quite drastically when $\alpha$ is set in an area $1 \leq \alpha \leq 2$, where it strongly increases its predictive performance ($pR^2$) at the cost of its document modelling performance (perplexity). This can be seen in the original paper \cite{chen_bpslda_2015}. We included $\alpha = 1$ in the robustness test range and BTR is still generally on par with BPsLDA in this specific case for low $K$ and does better for $K > 30$. Even when including $\alpha = 1$ in the robustness test range, BTR still outperforms BPsLDA and all other models across all hyperparameter settings, except $K=10$ in the Yelp dataset, where BTR is a close second.

\newpage
\subsection{Robustness Tests}
Robustness test across all topic models with LDA-like structure and Dirichlet hyperparameters for document-topic and word-topic distributions.

We assess the robustness of our findings to changes in the Dirichlet hyperparameters $\alpha$ and $\eta$. These hyperparameters act as priors on the topic-document distributions ($\bs{\beta}$) and word-topic distributions ($\bs{\theta}$), respectively. Table \ref{table:robustness_stats} shows the results. 

In terms of $pR^2$, BTR continues to perform best for all settings. We generally find that the BTR prediction performance is robust to hyperparameter changes. Evaluating the perplexity scores, we see more fluctuation across all models, which is unsurprising since those hyperparameter directly affect the generative topic modelling processes. BTR remains on par with its sLDA counterpart.
\begin{table} [H]
    \centering
    \caption{Sensitivity to hyperparameters $\alpha$ and $\beta$ (K = 20)}
    \label{table:robustness_stats}
    \begin{tabular}{cccccccc}
    \toprule
    & & \multicolumn{3}{c}{$\alpha$} & \multicolumn{3}{c}{$\eta$}   \\
    \textit{Metric} & \textit{Model} & 0.1 & 0.5 & 1 &  0.001 & 0.01 & 0.1 \\
    \midrule
    \multirow{4}{*}{Yelp $pR^2$} & LR-LDA & 0.473 & 0.530 & 0.550 & 0.316 & 0.530 & 0.521 \\
    & LR-sLDA & 0.558 & 0.564 & 0.559 & 0.562 & 0.564 & 0.568 \\
    & LR-BPsLDA & 0.602 & 0.597 & 0.608 & 0.607 & 0.597 & 0.608 \\
    & BTR & \textbf{0.611} & \textbf{0.615} & \textbf{0.613} & 0.\textbf{611} & \textbf{0.615} & \textbf{0.624} \\
    \midrule
    \multirow{3}{*}{Yelp perplexity} & LR-LDA & 1511 & 	1448 & 1445 & 1472 & 1447 & \textbf{1470} \\
    & LR-sLDA & 1497 & 1444 & \textbf{1431} & \textbf{1441} & 1444 & 1491 \\
    & BTR & \textbf{1490} & \textbf{1443} & 1441 & 1456 & \textbf{1443} & 1478 \\
    \midrule
    \multirow{4}{*}{Booking $pR^2$} & LR-LDA & 0.397 & 0.410 & 0.409 & 0.405 & 0.410 & 0.406 \\
    & LR-sLDA & 0.430 & 0.426 & 0.432 & 0.422 & 0.426 & 0.433 \\
    & LR-BPsLDA & 0.409 & 0.396 & \textbf{0.453} & 0.395 & 0.396 & 0.393\\
    & BTR & \textbf{0.451} & \textbf{0.447} & 0.452 & \textbf{0.443} & \textbf{0.447} & \textbf{0.455} \\
    \midrule
    \multirow{3}{*}{Booking perplexity} & LR-LDA & 515 & 498 & 514 & 505 & 498 & \textbf{512} \\
    & LR-sLDA & \textbf{502} & \textbf{491} & 504 & \textbf{484} & \textbf{491} & 516 \\
    & BTR &  503 & \textbf{491} & \textbf{503} & 489 & \textbf{491} & 515 \\
    \bottomrule
    \end{tabular}
\end{table}

\newpage
\subsubsection{Further Robustness Tests - Booking}
Table \ref{rob_pr_bk} provides an extended robustness test on the predictive performance of the benchmark topic models across hyperparameters. BTR continues to be the best performing model throughout. Table \ref{rob_pplxy_bk} summarises robustness tests in terms of perplexity scores. BTR achieves almost identical perplexity scores as sLDA whilst achieving higher $pR^2$ throughout.
\begin{table}[H]
  \centering
  \caption{Booking - $pR^2$ for different hyperparameter settings across topic benchmark models, best model in bold.}
    \resizebox{\linewidth}{5cm}{    \begin{tabular}{ccccccccccc}
    (K=10) & $\alpha$ = 0.1 & $\alpha$ = 0.5 & $\alpha$ = 1 & $\eta$ = 0.001 & $\eta$ = 0.01 & $\eta$ = 0.1 & a,b = (3,2) & a,b = (0,0) & a,b = (3,4) & a,b = (1.5,4) \\
    \midrule
    LR-LDA & 0.378 & 0.4   & 0.408 & 0.397 & 0.4   & 0.398 &       &       &       &  \\
    LR-sLDA & 0.42  & 0.416 & 0.403 & 0.401 & 0.416 & 0.422 &       &       &       &  \\
    LR-BPsLDA & 0.396 & 0.394 & 0.439 & 0.393 & 0.394 & 0.396 &       &       &       &  \\
    \textbf{BTR }  & 0.446 & 0.439 & 0.435 & 0.418 & 0.439 & \textbf{0.452} & 0.439 & 0.435 & 0.437 & 0.446 \\
    \midrule
          &       &       &       &       &       &       &       &       &       &  \\
    (K=20) & $\alpha$ = 0.1 & $\alpha$ = 0.5 & $\alpha$ = 1 & $\eta$ = 0.001 & $\eta$ = 0.01 & $\eta$ = 0.1 & a,b = (3,2) & a,b = (0,0) & a,b = (3,4) & a,b = (1.5,4) \\
    \midrule
    LR-LDA & 0.397 & 0.41  & 0.409 & 0.405 & 0.41  & 0.406 &       &       &       &  \\
    LR-sLDA & 0.43  & 0.426 & 0.432 & 0.422 & 0.426 & 0.433 &       &       &       &  \\
    LR-BPsLDA & 0.409 & 0.396 & 0.453 & 0.395 & 0.396 & 0.393 &       &       &       &  \\
    \textbf{BTR}   & 0.451 & 0.447 & 0.452 & 0.443 & 0.447 & \textbf{0.455} & 0.447 & 0.45  & 0.45  & 0.443 \\
    \midrule
          &       &       &       &       &       &       &       &       &       &  \\
    (K=30) & $\alpha$ = 0.1 & $\alpha$ = 0.5 & $\alpha$ = 1 & $\eta$ = 0.001 & $\eta$ = 0.01 & $\eta$ = 0.1 & a,b = (3,2) & a,b = (0,0) & a,b = (3,4) & a,b = (1.5,4) \\
    \midrule
    LR-LDA & 0.399 & 0.417 & 0.423 & 0.417 & 0.417 & 0.413 &       &       &       &  \\
    LR-sLDA & 0.434 & 0.43  & 0.428 & 0.417 & 0.43  & 0.427 &       &       &       &  \\
    LR-BPsLDA & 0.424 & 0.4   & 0.451 & 0.401 & 0.4   & 0.402 &       &       &       &  \\
    \textbf{BTR}   & 0.455 & 0.453 & 0.455 & 0.444 & 0.453 & \textbf{0.459} & 0.453 & 0.453 & 0.447 & 0.449 \\
    \midrule
          &       &       &       &       &       &       &       &       &       &  \\
    (K=50) & $\alpha$ = 0.1 & $\alpha$ = 0.5 & $\alpha$ = 1 & $\eta$ = 0.001 & $\eta$ = 0.01 & $\eta$ = 0.1 & a,b = (3,2) & a,b = (0,0) & a,b = (3,4) & a,b = (1.5,4) \\
    \midrule
    LR-LDA & 0.415 & 0.426 & 0.428 & 0.418 & 0.426 & 0.420 &       &       &       &  \\
    LR-sLDA & 0.434 & 0.432 & 0.43  & 0.429 & 0.432 & 0.436 &       &       &       &  \\
    \textbf{LR-BPsLDA} & \textbf{0.461} & 0.419 & 0.449 & 0.411 & 0.419 & 0.418 &       &       &       &  \\
    \textbf{BTR}   & \textbf{0.461} & 0.454 & 0.459 & 0.446 & 0.454 & 0.459 & 0.454 & 0.455 & 0.452 & 0.451 \\
    \bottomrule
    \multicolumn{11}{c}{\textit{Default model was $\alpha=0.5$, $\eta=0.01$, $a_0=3$, $b_0=2$.}} \\
    \multicolumn{11}{c}{\textit{Robustness tests kept all hyperparameters at default, then changing one hyperparameter at a time.}} \\
    \end{tabular}}%
  \label{rob_pr_bk}%
\end{table}%

\begin{table}[H]
  \centering
  \caption{Booking - perplexity scores for different hyperparameter settings across topic benchmark models, best model in bold.}
    \resizebox{\linewidth}{4.25cm}{\begin{tabular}{lrrrrrlrlll}
    K=10  & \multicolumn{1}{c}{$\alpha$ = 0.1} & \multicolumn{1}{c}{$\alpha$ = 0.5} & \multicolumn{1}{c}{$\alpha$ = 1} & \multicolumn{1}{c}{$\eta$ = 0.001} & \multicolumn{1}{c}{$\eta$ = 0.01} & $\eta$ = 0.1 & \multicolumn{1}{c}{a,b = (3,2)} & a,b = (0,0) & a,b = (3,4) & a,b = (1.5,4) \\
    \midrule
    LDA   & 562   & 538   & 539   & 539   & 538   & \multicolumn{1}{c}{545} &       &       &       &  \\
    LR-sLDA & 557   & 535   & 539   & 534   & 535   & \multicolumn{1}{c}{554} &       &       &       &  \\
    \textbf{BTR}   & 556   & 535   & 538   & \textbf{528}   & 535   & \multicolumn{1}{c}{548} & 535   & \multicolumn{1}{c}{535} & \multicolumn{1}{c}{537} & \multicolumn{1}{c}{536} \\
    \midrule
          &       &       &       &       &       &       &       &       &       &  \\
    K=20  & \multicolumn{1}{c}{$\alpha$ = 0.1} & \multicolumn{1}{c}{$\alpha$ = 0.5} & \multicolumn{1}{c}{$\alpha$ = 1} & \multicolumn{1}{c}{$\eta$ = 0.001} & \multicolumn{1}{c}{$\eta$ = 0.01} & $\eta$ = 0.1 & \multicolumn{1}{c}{a,b = (3,2)} & a,b = (0,0) & a,b = (3,4) & a,b = (1.5,4) \\
    \midrule
    LDA   & 515   & 498   & 514   & 505   & 498   & \multicolumn{1}{c}{512} &       &       &       &  \\
    \textbf{LR-sLDA} & 502   & 491   & 504   & \textbf{484}   & 491   & \multicolumn{1}{c}{516} &       &       &       &  \\
    BTR   & 503   & 490   & 503   & 489   & 490   & \multicolumn{1}{c}{515} & 490   & \multicolumn{1}{c}{491} & \multicolumn{1}{c}{490} & \multicolumn{1}{c}{491} \\
    \midrule
          &       &       &       &       &       &       &       &       &       &  \\
    K=30  & \multicolumn{1}{c}{$\alpha$ = 0.1} & \multicolumn{1}{c}{$\alpha$ = 0.5} & \multicolumn{1}{c}{$\alpha$ = 1} & \multicolumn{1}{c}{$\eta$ = 0.001} & \multicolumn{1}{c}{$\eta$ = 0.01} & $\eta$ = 0.1 & \multicolumn{1}{c}{a,b = (3,2)} & a,b = (0,0) & a,b = (3,4) & a,b = (1.5,4) \\
    \midrule
    LDA   & 479   & 476   & 502   & 484   & 476   & \multicolumn{1}{c}{492} &       &       &       &  \\
    \textbf{LR-sLDA} & 471   & 463   & 486   & \textbf{454 }  & 463   & \multicolumn{1}{c}{499} &       &       &       &  \\
   BTR   & 470   & 463   & 483   & 457   & 463   & \multicolumn{1}{c}{500} & 463   & \multicolumn{1}{c}{463} & \multicolumn{1}{c}{463} & \multicolumn{1}{c}{463} \\
    \midrule
          &       &       &       &       &       &       &       &       &       &  \\
    K=50  & \multicolumn{1}{c}{$\alpha$ = 0.1} & \multicolumn{1}{c}{$\alpha$ = 0.5} & \multicolumn{1}{c}{$\alpha$ = 1} & \multicolumn{1}{c}{$\eta$ = 0.001} & \multicolumn{1}{c}{$\eta$ = 0.01} & $\eta$ = 0.1 & \multicolumn{1}{c}{a,b = (3,2)} & a,b = (0,0) & a,b = (3,4) & a,b = (1.5,4) \\
    \midrule
    LDA   & 442   & 454   & 494   & 460   & 454   & \multicolumn{1}{c}{476} &       &       &       &  \\
    \textbf{LR-sLDA} & 431   & 436   & 466   & \textbf{421}   & 436   & \multicolumn{1}{c}{492} &       &       &       &  \\
    BTR   & 430     & 437   & 467   & 423   & 437   & 
    \multicolumn{1}{c}{492} & 437   & \multicolumn{1}{c}{436} & \multicolumn{1}{c}{439} & \multicolumn{1}{c}{437}  \\
    \bottomrule
    \multicolumn{11}{c}{\textit{Default model was $\alpha=0.5$, $\eta=0.01$, $a_0=3$, $b_0=2$.}} \\
    \multicolumn{11}{c}{\textit{Robustness tests kept all hyperparameters at default, then changing one hyperparameter at a time.}} \\
    \end{tabular}}%
  \label{rob_pplxy_bk}%
\end{table}%

\subsubsection{Further Robustness Tests - Yelp}
Table \ref{app:rob_pr_yp} provides an extended robustness test on the predictive performance of the benchmark topic models across hyperparameters. BTR continues to be the best performing model throughout, apart from the K=10 case, where it is a close second. 
Table \ref{app_pplxy_yp} summarises robustness tests in terms of perplexity scores. BTR achieves almost identical perplexity scores as sLDA whilst achieving higher $pR^2$ throughout.
\begin{table}[H]
  \centering
  \caption{Yelp - $pR^2$ for different hyperparameter settings across topic benchmark models, best model in bold.}
        \resizebox{\linewidth}{5cm}{\begin{tabular}{ccccccccccc}
    K=10 & $\alpha$ = 0.1 & $\alpha$ = 0.5 & $\alpha$ = 1 & $\eta$ = 0.001 & $\eta$ = 0.01 & $\eta$ = 0.1 & \multicolumn{1}{c}{a,b = (4,2)} & a,b = (0,0) & a,b = (3,4) & a,b = (1.5,4) \\
    \midrule
    LR-LDA & 0.476 & 0.498 & 0.515 & 0.503 & 0.498 & 0.49  &       &       &       &  \\
    LR-sLDA & 0.523 & 0.533 & 0.539 & 0.52  & 0.533 & 0.527 &       &       &       &  \\
    \textbf{LR-BPsLDA} & 0.596 & 0.593 & \textbf{0.606} & 0.595 & 0.593 & 0.592 &       &       &       &  \\
    BTR   & 0.592 & 0.586 & 0.593 & 0.575 & 0.586 & 0.596 & 0.586 & 0.588 & 0.578 & 0.59 \\
    \midrule
          &       &       &       &       &       &       &       &       &       &  \\
    K=20 & $\alpha$ = 0.1 & $\alpha$ = 0.5 & $\alpha$ = 1 & $\eta$ = 0.001 & $\eta$ = 0.01 & $\eta$ = 0.1 & \multicolumn{1}{c}{a,b = (4,2)} & a,b = (0,0) & a,b = (3,4) & a,b = (1.5,4) \\
    \midrule
    LR-LDA & 0.473 & 0.53  & 0.55  & 0.483 & 0.53  & 0.521 &       &       &       &  \\
    LR-sLDA & 0.558 & 0.564 & 0.559 & 0.562 & 0.564 & 0.568 &       &       &       &  \\
    LR-BPsLDA & 0.602 & 0.597 & 0.608 & 0.607 & 0.597 & 0.608 &       &       &       &  \\
    \textbf{BTR}   & 0.611 & 0.615 & 0.613 & 0.611 & 0.615 & \textbf{0.624} & 0.615 & 0.62  & 0.593 & 0.621 \\
    \midrule
          &       &       &       &       &       &       &       &       &       &  \\
    K=30 & $\alpha$ = 0.1 & $\alpha$ = 0.5 & $\alpha$ = 1 & $\eta$ = 0.001 & $\eta$ = 0.01 & $\eta$ = 0.1 & \multicolumn{1}{c}{a,b = (4,2)} & a,b = (0,0) & a,b = (3,4) & a,b = (1.5,4) \\
    \midrule
    LR-LDA & 0.499 & 0.561 & 0.563 & 0.547 & 0.561 & 0.565 &       &       &       &  \\
    LR-sLDA & 0.565 & 0.567 & 0.563 & 0.567 & 0.567 & 0.56  &       &       &       &  \\
    LR-BPsLDA & 0.609 & 0.597 & 0.607 & 0.599 & 0.597 & 0.599 &       &       &       &  \\
    \textbf{BTR}   & 0.624 & \textbf{0.627} & 0.612 & 0.608 &\textbf{ 0.627} & 0.622 & \textbf{0.627} & 0.623 & \textbf{0.627 }& 0.626 \\
    \midrule
          &       &       &       &       &       &       &       &       &       &  \\
    K=50 & $\alpha$ = 0.1 & $\alpha$ = 0.5 & $\alpha$ = 1 & $\eta$ = 0.001 & $\eta$ = 0.01 & $\eta$ = 0.1 & \multicolumn{1}{c}{a,b = (4,2)} & a,b = (0,0) & a,b = (3,4) & a,b = (1.5,4) \\
    \midrule
    LR-LDA & 0.523 & 0.586 & 0.591 & 0.571 & 0.586 & 0.582 &       &       &       &  \\
    LR-sLDA & 0.573 & 0.571 & 0.564 & 0.556 & 0.571 & 0.573 &       &       &       &  \\
    LR-BPsLDA & 0.612 & 0.603 & 0.606 & 0.604 & 0.603 & 0.604 &       &       &       &  \\
    \textbf{BTR}   & \textbf{0.632} & 0.630  & 0.623 & 0.621 & 0.630  & \textbf{0.632} & 0.630  & 0.629 & 0.629 & 0.628 \\
    \bottomrule
    \end{tabular}}%
  \label{app:rob_pr_yp}%
\end{table}%
\begin{table}[H]
  \centering
  \caption{Yelp - perplexity scores for different hyperparameter settings across topic benchmark models, best model in bold.}
    \resizebox{\linewidth}{4cm}{\begin{tabular}{ccccccccccc}
    K=10 & $\alpha$ = 0.1 & $\alpha$ = 0.5 & $\alpha$ = 1 & $\eta$ = 0.001 & $\eta$ = 0.01 & $\eta$ = 0.1 & \multicolumn{1}{c}{a,b = (4,2)} & a,b = (0,0) & a,b = (3,4) & a,b = (1.5,4) \\
    \midrule
    LDA   & 1586  & 1544  & 1532  & 1557  & 1544  & 1552  & 1544  &       &       &  \\
    \textbf{LR-sLDA} & 1583  & 1544  & \textbf{1530} & 1561  & 1544  & 1554  & 1544  &       &       &  \\
    BTR  & 1588  & 1540  & 1534  & 1565  & 1540  & 1546  & 1540  & 1539 & 1548 & 1547 \\
    \midrule
          &       &       &       &       &       &       &       &       &       &  \\
    K=20 & $\alpha$ = 0.1 & $\alpha$ = 0.5 & $\alpha$ = 1 & $\eta$ = 0.001 & $\eta$ = 0.01 & $\eta$ = 0.1 & \multicolumn{1}{c}{a,b = (4,2)} & a,b = (0,0) & a,b = (3,4) & a,b = (1.5,4) \\
    \midrule
    LDA   & 1511  & 1447  & 1445  & 1472  & 1447  & 1469  & 1447  &       &       &  \\
    \textbf{LR-sLDA} & 1497  & 1444  & \textbf{1431} & 1441  & 1444  & 1491  & 1444  &       &       &  \\
    BTR  & 1490  & 1443  & 1441  & 1456  & 1443  & 1478  & 1443  & 1443 & 1445 & 1441 \\
    \midrule
          &       &       &       &       &       &       &       &       &       &  \\
    K=30 & $\alpha$ = 0.1 & $\alpha$ = 0.5 & $\alpha$ = 1 & $\eta$ = 0.001 & $\eta$ = 0.01 & $\eta$ = 0.1 & \multicolumn{1}{c}{a,b = (4,2)} & a,b = (0,0) & a,b = (3,4) & a,b = (1.5,4) \\
    \midrule
    LDA   & 1434  & 1388  & 1390  & 1412  & 1388  & 1415  & 1388  &       &       &  \\
    LR-sLDA & 1436  & 1382  & 1383  & 1395  & 1382  & 1442  & 1382  &       &       &  \\
    \textbf{BTR} & 1434  & \textbf{1379} & 1385  & 1390  & \textbf{1379} & 1448  & \textbf{1379} & 1378 & 1389 & 1379 \\
    \midrule
          &       &       &       &       &       &       &       &       &       &  \\
    K=50 & $\alpha$ = 0.1 & $\alpha$ = 0.5 & $\alpha$ = 1 & $\eta$ = 0.001 & $\eta$ = 0.01 & eta = 0.1 & \multicolumn{1}{c}{a,b = (4,2)} & a,b = (0,0) & a,b = (3,4) & a,b = (1.5,4) \\
    \midrule
    LDA   & 1352  & 1306  & 1325  & 1334  & 1306  & 1356  & 1306  &       &       &  \\
    LR-sLDA & 1349  & 1294  & 1310  & 1309  & 1294  & 1404  & 1294  &       &       &  \\
    \textbf{BTR} & 1338  & \textbf{1291} & 1303  & 1288  & \textbf{1291} & 1405  & \textbf{1291} & 1293 & 1294 & 1292 \\
    \bottomrule
    \end{tabular}}%
  \label{app_pplxy_yp}%
\end{table}%

\subsection{Estimated Topics}
\label{section:app_topics}
The below tables are an extended version of the corresponding table in the paper. The show the top $3$ negative and positive topics for $K=[10,30,100]$. Inspecting the top words in each of these topics compared with its regression coefficient, BTR models highly interpretable topics - at least as interpretable as LDA or sLDA. At the same time BTR achieves substantially better prediction performances throughout all model specifications (see previous section).
\twocolumn
\begin{table}[t!]
    \centering
    \small
    \caption{Top 3 positive and negative topics for \textit{Yelp} (K = 10)}
    \label{tab:example_topics}
    \resizebox{8cm}{3.33cm}{\begin{tabular}{@{\extracolsep{-2pt}} lcccccc}
    \toprule
     & Pos1 & Pos2 & Pos3 & Neg3 & Neg2 & Neg1  \\ 
    \midrule
    \multirow{6}{*}{BTR topics} & food & restaur & time & store & food & order \\
    & great & dish & hair & locat & chicken & us \\
    & place & lobster & back & like & good & ask \\
    & servic & menu & work & can & order & servic \\
    & friend & food & will & price & rice & wait \\
    & love & order & day & go & dish & food \\
    \midrule
    BTR regr. weights & 4.3 & 1.7 & 1.5 & -0.1 & -0.5 & -8.8 \\
    \bottomrule
    \multirow{6}{*}{sLDA topics} & food & time & locat & coffe & us & like \\
    & place & hair & store & tea & order & place \\
    & great & back & can & tri & ask & go \\
    & servic & work & find & place & servic & much \\
    & good & will & place & ice & tabl & im \\
    & time & day & staff & cream & time & realli \\
    \midrule 
    sLDA regr. weights & 2.7 & 1.7 & 1.2 & 0.1 & -3.7 & -4.5 \\
    \bottomrule
    \multirow{6}{*}{LDA topics} & food & place & place & store & fri & order \\
    & great & coffe & great & like & burger & us \\
    & servic & good & good & locat & order & food \\
    & restaur & tri & friend & can & like & servic \\
    & dish & tea & bar & find & good & time \\
    & menu & great & drink & go & chees & ask \\
    \midrule 
    LDA regr. weights & 1.2 & 0.7 & 0.5 & -0.1 & -0.4 & -2.6 \\
    \bottomrule
    \end{tabular}}
\end{table}

\begin{table}[b!]
    \centering
    \small
    \caption{Top 3 positive and negative topics for \textit{Yelp} (K = 30)}
    \label{tab:example_topics}
    \resizebox{8cm}{3.33cm}{\begin{tabular}{@{\extracolsep{-2pt}} lcccccc}
    \toprule
     & Pos1 & Pos2 & Pos3 & Neg3 & Neg2 & Neg1  \\ 
    \midrule
    \multirow{6}{*}{BTR topics} & best & great & restaur & us & ask & like \\
    & plac & friend & menu & order & said & disappoint \\
    & alway & servic & dish & tabl & custom & better \\
    & love & staff & wine & food & told & tast \\
    & ever & recommend & steak & server & never & noth \\
    & toronto & amaz & perfect & came & say & bad \\
    \midrule
    BTR regr. weights & 6.9 & 6.0 & 2.1 & -3.9 & -8.4 & -13.3 \\
    \bottomrule
    \multirow{6}{*}{sLDA topics} & great & time & im & seem & ask & like \\
    & love & alway & review & like & never & food \\
    & amaz & go & place & much & custom & good \\
    & recommend & year & star & make & said & place \\
    & servic & never & go & think & servic & tast \\
    & friend & everi & give & thing & told & better \\
    \midrule 
    sLDA regr. weights & 3.7 & 3.1 & 3.1 & -2.3 & -6.4 & -7.1 \\
    \bottomrule
    \multirow{6}{*}{LDA topics} & great & toronto & restaur & us & ask & like \\
    & friend & visit & menu & tabl & custom & tast \\
    & love & make & dish & order & said & disappoint \\
    & amaz & love & wine & food & servic & better \\
    & place & made & dessert & came & told & bad \\
    & servic & best & dinner & server & manag & noth \\
    \midrule 
    LDA regr. weights & 3.0 & 1.6 & 1.3 & -1.2 & -4.9 & -8.3 \\
    \bottomrule
    \end{tabular}}
\end{table}

\begin{table}[t!]
    \centering
    \small
    \caption{Top 3 positive and negative topics for \textit{Yelp} (K = 100)}
    \label{tab:example_topics}
    \resizebox{8cm}{3.33cm}{\begin{tabular}{@{\extracolsep{-2pt}} lcccccc}
    \toprule
     & Pos1 & Pos2 & Pos3 & Neg3 & Neg2 & Neg1  \\ 
    \midrule
    \multirow{6}{*}{BTR topics} & love & definit & best & custom & never & disappoint \\
    & delici & ever & amaz & ask & worst & tast \\
    & definit & toronto & everi & said & ever & bland \\
    & perfect & citi & friend & manag & money & dri \\
    & tri & far & free & rude & bad & better \\
    & super & amaz & alway & servic & terribl & lack \\
    \midrule 
    BTR regr. weights & 5.9 & 5.2 & 5.0 & -8.4 & -12.5 & -14.5 \\
    \bottomrule
    \multirow{6}{*}{sLDA topics} & alway & will & amaz & ask & tast & disappoint \\
    & time & definit & definit & said & like & bad \\
    & usual & servic & love & told & felt & cold \\
    & come & friend & great & back & disappoint & worst \\
    & never & return & place & went & better & dri \\
    & everi & back & everyth & want & wasnt & lack \\
    \midrule 
    sLDA regr. weights & 4.1 & 4.0 & 3.9 & -7.0 & -8.0 & -11.3 \\
    \bottomrule
    \multirow{6}{*}{LDA topics} & love & best & experi & money & never & tast \\
    & amaz & toronto & made & go & bad & like \\
    & delici & ever & make & will & ever & disappoint \\
    & place & citi & feel & never & worst & meat \\
    & absolut & far & first & pay & terribl & bland \\
    & super & visit & felt & spend & experi & dri \\
    \midrule 
    LDA regr. weights & 5.4 & 4.6 & 3.8 & -5.9 & -10.8 & -10.9 \\
    \bottomrule
    \end{tabular}}
\end{table}

\begin{table}[t!]
    \centering
    \small
    \caption{Top 3 positive and negative topics for \textit{Booking} (K = 10)}
    \label{tab:example_topics}
    \resizebox{8cm}{3.33cm}{\begin{tabular}{@{\extracolsep{-2pt}} lcccccc}
    \toprule
     & Pos1 & Pos2 & Pos3 & Neg3 & Neg2 & Neg1  \\ 
    \midrule
    \multirow{6}{*}{BTR topics} & hotel & room & room & room & check & hotel \\
    & stay & locat & great & bed & book & room \\
    & staff & staff & love & shower & room & locat \\
    & would & good & hotel & bathroom & us & small \\
    & help & clean & view & small & hotel & good \\
    & everyth & comfort & bar & clean & arriv & price \\
    \midrule
    BTR regr. weights & 2.8 & 1.7 & 1.5 & -1.0 & -1.1 & -5.7 \\
    \bottomrule
    \multirow{6}{*}{sLDA topics} & hotel & room & room & room & room & room \\
    & stay & good & great & bed & night & hotel \\
    & staff & locat & love & bathroom & window & locat \\
    & would & staff & hotel & shower & work & small \\
    & help & clean & view & small & floor & staff \\
    & like & breakfast & nice & comfort & air & posit \\
    \midrule 
    sLDA regr. weights & 2.4 & 1.3 & 1.3 & -0.4 & -0.6 & -5.6 \\
    \bottomrule
    \multirow{6}{*}{LDA topics} & hotel & hotel & neg & check & room & room \\
    & stay & great & staff & room & shower & hotel \\
    & staff & love & locat & book & bathroom & good \\
    & help & room & friendli & hotel & work & locat \\
    & would & view & great & us & bed & breakfast \\
    & noth & locat & help & time & air & price \\
    \midrule 
    LDA regr. weights & 1.3 & 1.2 & 1.0 & -1.3 & -1.4 & -2.0 \\
    \bottomrule
    \end{tabular}}
\end{table}

\begin{table}[b!]
    \centering
    \small
    \caption{Top 3 positive and negative topics for \textit{Booking} (K = 30)}
    \label{tab:example_topics}
    \resizebox{8cm}{3.33cm}{\begin{tabular}{@{\extracolsep{-2pt}} lcccccc}
    \toprule
     & Pos1 & Pos2 & Pos3 & Neg3 & Neg2 & Neg1  \\ 
    \midrule
    \multirow{6}{*}{BTR topics} &  us & stay & room & room & ask & room \\
    & staff & would & locat & small & us & hotel \\
    & made & hotel & great & bed & day & old \\
    & upgrad & staff & staff & size & recept & poor \\
    & stay & love & bit & locat & call & star \\
    & welcom & recommend & littl & bathroom & back & bad \\
    \midrule 
    BTR regr. weights & 2.7 & 2.5 & 2.3 & -2.5 & -3.0 & -9.0 \\
    \bottomrule
    \multirow{6}{*}{sLDA topics} & staff & hotel & us & book & room & hotel \\
    & friendli & love & upgrad & charg & need & room \\
    & great & beauti & staff & hotel & locat & bad \\
    & help & decor & room & pay & old & star \\
    & locat & modern & stay & check & look & poor \\
    & neg & great & love & day & smell & posit \\
    \midrule 
    sLDA regr. weights & 2.1 & 1.8 & 1.7 & -1.4 & -2.1 & -9.7 \\
    \bottomrule
    \multirow{6}{*}{LDA topics} & stay & stay & hotel & us & room & hotel \\
    & hotel & hotel & love & ask & locat & like \\
    & made & would & beauti & one & good & star \\
    & like & recommend & great & recept & need & realli \\
    & feel & definit & decor & day & old & much \\
    & realli & love & staff & call & valu & best \\
    \midrule 
    LDA regr. weights & 2.4 & 2.1 & 1.9 & -2.5 & -2.7 & -3.4 \\
    \bottomrule
    \end{tabular}}
\end{table}

\begin{table}[t!]
    \centering
    \small
    \caption{Top 3 positive and negative topics for \textit{Booking} (K = 100)}
    \label{tab:example_topics}
    \resizebox{8cm}{3.33cm}{\begin{tabular}{@{\extracolsep{-2pt}} lcccccc}
    \toprule
     & Pos1 & Pos2 & Pos3 & Neg3 & Neg2 & Neg1  \\ 
    \midrule
    \multirow{6}{*}{BTR topics} & staff & hotel & love & old & room & poor \\
    & help & wonder & great & look & small & posit \\
    & friendli & beauti & staff & carpet & tini & servic \\
    & excel & love & littl & tire & bathroom & bad \\
    & especi & experi & fab & furnitur & noisi & never \\
    & wonder & fabul & especi & need & far & rude \\
    \midrule 
    BTR regr. weights & 4.3 & 4.1 & 3.3 & -6.1 & -7.3 & -14.2 \\
    \bottomrule
    \multirow{6}{*}{sLDA topics} & love & room & great & old & hotel & bad \\
    & beauti & small & locat & dirti & star & poor \\
    & amaz & posit & neg & bathroom & expect & recept \\
    & fantast & size & perfect & carpet & rate & posit \\
    & fabul & bit & awesom & wall & thi & even \\
    & wonder & expect & super & look & basic & never \\
    \midrule 
    sLDA regr. weights & 3.7 & 3.6 & 2.8 & -5.8 & -6.0 & -14.3 \\
    \bottomrule
    \multirow{6}{*}{LDA topics} & love & great & bit & hotel & old & recept \\
    & amaz & locat & littl & star & dirti & manag \\
    & everyth & neg & nice & rate & carpet & rude \\
    & noth & staff & locat & expect & look & receptionist \\
    & perfect & awesom & breakfast & disappoint & wall & check \\
    & absolut & perfect & good & thi & furnitur & guest \\
    \midrule 
    LDA regr. weights & 3.6 & 3.1 & 2.8 & -5.3 & -6.6 & -7.6 \\
    \bottomrule
    \end{tabular}}
\end{table}

\onecolumn

\subsection{Computation Times}

Table \ref{tab:comp_time} shows the time taken for 100 E-step iterations on a single 2.8GHz processor on the Booking data and 300-400 seconds on the Yelp data. We found that 100 E-step iterations is typically sufficient for the best performance and the model typically converges after between 10-25 EM iterations. A typical 30 topic model on Yelp data thus took around 1 hour to converge, and around 20 minutes for Booking. Computation time scales roughly linearly in the number of topics and total number of words across all documents. This is because the evaluation of the $K$-dimensional multinomial distribution for each $z_{d,n}$ (equation \eqref{sLDA_cov_sampling_form}) is the principle computational challenge.

\begin{table}[H]
    \centering
    \caption{Computational time}
    \label{tab:comp_time}
    \begin{tabular}{ccc}
        \toprule
         Dataset & K  & 100 E-step iters \\
         \midrule 
         \multirow{5}{*}{Yelp} & 10  &  50s \\
         & 20 & 110s \\
         & 30 & 200s \\
         & 50 & 320s \\
         & 100 & 740s \\
         \bottomrule
         \multirow{5}{*}{Booking}  & 10  & 18s \\
         & 20 & 33s \\
         & 30 & 50s \\
         & 50 & 79s \\
         & 100 & 200s \\
         \bottomrule
    \end{tabular}
    \caption*{\textit{Note}: Yelp data has roughly 3 times as many words as Booking.com data}
\end{table}

\end{document}